\documentclass[pdflatex,sn-mathphys-num]{sn-jnl}

\usepackage[utf8]{inputenc}
\usepackage{amsmath,amssymb,amsfonts}
\usepackage{amsthm}
\usepackage{mathrsfs}
\usepackage{xcolor}
\usepackage{textcomp}

\usepackage{booktabs}

\usepackage{array}
\usepackage{booktabs}
\usepackage{multirow}
\usepackage{hhline}
\usepackage{lmodern}

\usepackage{algorithm}
\usepackage{algorithmicx}
\usepackage{algpseudocode}
\usepackage{listings}

\usepackage{graphicx}
\usepackage{float}
\usepackage{caption}
\usepackage{svg}
\usepackage[labelformat=parens]{subcaption}  

\usepackage{tikz, pgfplots, pgf-pie}
\usetikzlibrary{positioning}
\usepackage{booktabs}\let\cline\cmidrule

\usepackage[title]{appendix}
\usepackage{manyfoot}
\usepackage{hyperref}
\usepackage{url}
\usepackage[backend=biber, style=numeric-comp, sorting=none]{biblatex}
\usepackage[utf8]{inputenc}
\usepackage{pifont}
\usepackage{newunicodechar}
\newunicodechar{✓}{\ding{51}}
\usepackage{changepage}
\usepackage{array}
\usepackage{siunitx}
\usepackage{makecell}  
\usepackage{amsmath} 
\usepackage{enumitem}

\usepackage{appendix}
\usepackage{longtable}
\usepackage{booktabs}
\usepackage{xcolor}
\usepackage{array}
\usepackage{framed}
\usepackage{bbding}

\usepackage{tikz}

\def\halfcheckmark{\tikz\draw[scale=0.25,fill=black](0,.35) -- (.25,0) -- (1,.7) -- (.25,.15) -- cycle (0.95,0.2) -- (0.77,0.2)  -- (0.6,0.7) -- cycle;}

\newcommand{\xmark}{\ding{55}} 

\theoremstyle{thmstyleone}%
\theoremstyle{thmstyletwo}%

\theoremstyle{thmstylethree}%

\def\ie{{\em i.e.}}
\def\eg{{\em e.g.}}

\begin{document}
\title[Fit for Purpose? Deepfake Detection in the Real World]{Fit for Purpose? Deepfake Detection in the Real World}




\author[1]{\fnm{Guangyu} \sur{Lin}}\email{lin2285@purdue.edu}
\equalcont{These authors contributed equally to this work.}

\author[1]{\fnm{Li} \sur{Lin}}\email{lin1785@purdue.edu}
\equalcont{These authors contributed equally to this work.}

\author[2]{\fnm{Christina P.} \sur{Walker}}\email{walke667@purdue.edu}

\author[2]{\fnm{Daniel S.} \sur{Schiff}}\email{dschiff@purdue.edu}

\author*[1]{\fnm{Shu} \sur{Hu}}\email{hu968@purdue.edu}

\affil*[1]{\orgdiv{School of Applied and Creative Computing}, \orgname{Purdue University}, \orgaddress{\city{West Lafayette}, \postcode{47907}, \state{IN}, \country{USA}}}

\affil[2]{\orgdiv{Department of Political Science}, \orgname{Purdue University}, \orgaddress{\city{West Lafayette}, \postcode{47907}, \state{IN}, \country{USA}}}

\abstract{The rapid proliferation of AI-generated content, driven by advances in generative adversarial networks, diffusion models, and multimodal large language models, has made the creation and dissemination of synthetic media effortless, heightening the risks of misinformation, particularly political deepfakes that distort truth and undermine trust in political institutions. In turn, governments, research institutions, and industry have strongly promoted deepfake detection initiatives as solutions. Yet, most existing models are trained and validated on synthetic, laboratory-controlled datasets, limiting their generalizability to the kinds of real-world political deepfakes circulating on social platforms that affect the public. 
In this work, we introduce the first systematic benchmark based on the Political Deepfakes Incident Database—a curated collection of real-world political deepfakes shared on social media since 2018. Our study includes a systematic evaluation of state-of-the-art deepfake detectors across academia, government, and industry. 
We find that the detectors from academia and government perform relatively poorly. 
While paid detection tools achieve relatively higher performance than free-access models, all evaluated detectors struggle to generalize effectively to authentic political deepfakes, and are vulnerable to simple manipulations, especially in the video domain. Results urge the need for politically contextualized deepfake detection frameworks to better safeguard the public in real-world settings.
}




\maketitle

\section{Introduction}\label{sec1}


    
    

    


AI-generated content (AIGC) has become increasingly prevalent in daily life, reshaping online communication \cite{krepsAllNewsThats2022, goldsteinHowDisinformationEvolved2021, worldeconomicforumTheseAre32024}. Advances in generative adversarial networks (GANs), diffusion models (DMs), and multimodal large language models (LLMs) have lowered the barriers to producing synthetic media, including fake images and video, which are now widespread across platforms like Twitter/X, Facebook, Instagram, and Reddit \cite{chesneyDeepFakesLooming2019}. However, the rapid proliferation of AIGC has introduced considerable risks \cite{laineUnderstandingEthicsGenerative2025, coeckelberghLLMsTruthDemocracy2025, barnettEthicalImplicationsGenerative2023, weidingerEthicalSocialRisks2021}.
Synthetic content, particularly \textit{deepfakes},\footnote{Deepfake is a portmanteau of “deep learning” and “fake.” It refers to AI-generated or digitally manipulated media (such as images or videos), produced using deep neural networks that appear highly realistic. Deepfakes often depict individuals engaging in actions or speaking words they never actually performed, thereby creating a deceptive yet convincing imitation of reality.} can blur the boundaries between authentic and fabricated information, erode trust in media and democratic institutions, fuel polarization, and enable reputational attacks or false claims of manipulation \cite{hawkinsDeepfakesDemandRise2025, wackGenerativePropagandaEvidence2025, loewensteinMakeAmericaFake2024, altayPeopleAreSkeptical2024}. Among the most concerning applications of deepfakes are their use in \textbf{politics} \cite{trifonovaMisinformationFraudStereotyping2024, birdTypologyRisksGenerative2023, bariachHarmsTaxonomyAI2024}. High-profile political deepfakes have appeared globally: from a spoof video of former U.S. President Barack Obama, distorted footage of Indian Prime Minister Narendra Modi, to fake election ads in countries from Australia to Slovakia  \cite{walkerMergingAIIncidents2024, linzerAdaptationInnovationCivic2025, denadalDeepfakeHypeAI2024, barnesDisinformationDeepfakesPlayed2024}. Since the popularization of diffusion and LLMs in 2022, the volume and accessibility of this type of content have surged, prompting calls for regulation, watermarking, and deepfake detection \cite{rainieAIPolitics242024, worldeconomicforumTheseAre32024, srinivasanDetectingAIFingerprints2024, whitehouseExecutiveOrderSafe2023, hineNewDeepfakeRegulations2022, faircloughVotersOverwhelminglyBelieve2024, leibowiczPrinciplesPracticesLessons2024, partnershiponaiPAIsResponsiblePractices2023, vasseiTransparencyWeTrust2024, lisinskaWhyWatermarkingText2024, harrisMetasAIWatermarking}. 

In response, there has been a concerted effort to develop deepfake detection models \cite{chollet2017xception,lin2024preserving,qian2020thinking}. 
The U.S. NIST Center for AI Standards and Innovation (formerly the AI Safety Institute), for instance, has emphasized the urgent need to detect and mitigate synthetic media, a priority echoed in several U.S. federal executive orders, the America’s AI Action Plan, and initiatives from multiple international AI safety institutes, regulatory bodies, legislative frameworks, and global competitions. \cite{whitehouseExecutiveOrderSafe2023, AI2025plan, NIST2024FrontierResearch, NIST2024ReducingRisksSyntheticContent, dolhansky2020deepfake, competition}. However, existing models are typically trained and evaluated on synthetic datasets, such as FaceForensics++ \cite{rossler2019faceforensics++}, Celeb-DF \cite{li2020celeb}, and DFDC \cite{dolhansky2020deepfake}, which are \textbf{created under controlled laboratory conditions}. While effective in their domains, these detection tools and benchmarks may fail to reflect the complexity and diversity of synthetic media encountered on social media platforms, which tend to defy neat categorization: varying in AI techniques used to create them, resolution, content structure, composition, and salience.

\begin{figure}[t]
    \centering
        \includegraphics[width=1\linewidth]{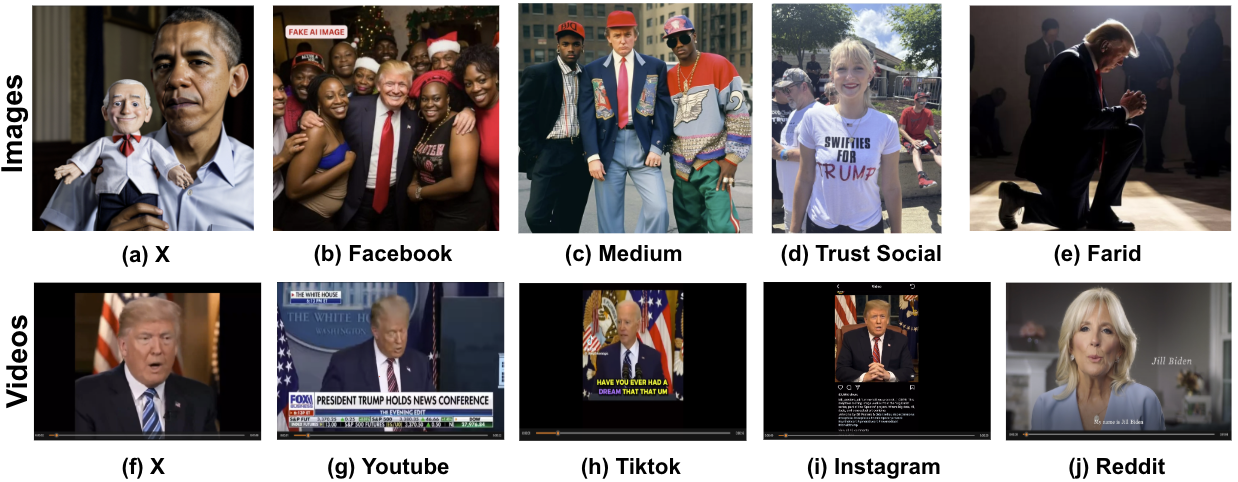}
        \caption{\emph{Political deepfake samples from different sources in the PDID.}}
    \label{fig:data_samples}
\end{figure}

For instance, many detection tools are focused squarely on detecting deepfakes rather than simpler manipulations, and emphasize detection of inauthentic renderings of humans and human faces in particular \cite{verdolivaMediaForensicsDeepFakes2020, tolosanaDeepfakesSurveyFace2020}. However, in real-world settings where individuals are most often exposed to potential misinformation, like social media, political deepfakes often contain multiple people, altered context such as background settings, mixed synthetic and real elements, and a range of image styles---cartoonish memes of a politician, burning of a flag, or an overhead view of a protest. Further, many instances of politically misleading images and videos do not meet the strict definition of a deepfake. For example, cheapfakes and shallowfakes--simpler forms of misinformation based on Photoshop-style tools, splicing, cropping, or other simple editing--can be equally persuasive and damaging, yet are often \textbf{overlooked} in detection efforts \cite{hameleersHowPersuasiveAre2024, hameleersCheapDeepManipulation2024, ioannouDeepfakesCheapfakesTwitter}. This oversight can lead to unintended but serious consequences. Social media platforms that adopt detection tools or watermarking and labeling schema like C2PA's Condent Credentials \cite{c2pa_spec}, for example, could risk falsely reassuring users that unmarked content is trustworthy, even when it is misleading or manipulated. This presents a challenge for the evaluation and deployment of detection tools: \textit{\textbf{how generalizable and robust are current state-of-the-art deepfake detectors to the kinds of content that actually circulate in high-stakes political contexts?}}

Recent efforts have begun to address these generalization concerns. For instance, Deepfake-Eval-2024 \cite{chandraDeepfakeEval2024MultiModalIntheWild2025} collects real-world examples from the web, while MAVOS-DD \cite{croitoruMAVOSDDMultilingualAudioVideo2025} introduces synthetically generated multilingual, multimodal benchmarks. Yet, despite important contributions, these benchmarks have significant limitations: they lack systematic coverage of politically salient incidents, still rely primarily on synthetic data, and fail to evaluate robustness under common post-processing operations such as blurring and compression  \cite{luImpactVideoProcessing2023}. \textit{\textbf{No existing benchmarking effort explicitly centers on real-world political deepfakes that are publicly and organically circulated.}}

In this work, we perform the \textbf{first} systematic benchmarking exercise against our current developed \textbf{Political Deepfakes Incident Database} (PDID) \cite{walkerMergingAIIncidents2024}, which curates politically salient deepfake incidents primarily from a United States context (including false allegations of deepfakes) sourced directly from real-world social media dissemination spanning from 2018 to the present (see samples in Fig. \ref{fig:data_samples}). 
Specifically, (1) we first select images and videos from the PDID database spanning 2018 to September 2025 and apply a human-in-the-loop, two-stage annotation process to assign the most accurate labels (fake or real) to each sample.
(2) We then analyze the dataset’s distribution in terms of source platform, resolution, publication year, and duration (for videos), and compare it with existing lab-based deepfake videos, GAN-generated images, and DM-generated images through a frequency-domain analysis to highlight distinctive characteristics of real-world fakes.
(3) Finally, we categorize existing detectors into two groups based on whether they incorporate or neglect large vision-language models (LVLMs): LVLM-aware detectors and LVLM-agnostic detectors. The LVLM-agnostic group is further divided into white-box detectors (from academia and government with accessible source code) and black-box detectors (from industry). We then systematically evaluate their performance on our curated PDID dataset.

We find that (1) the fake samples in the PDID display distinct spectral characteristics compared to lab-generated fakes, while the real samples in the PDID also differ substantially from those used in laboratory-generated datasets, revealing their \textbf{real-world complexity and variability} compared to synthetic datasets; (2) {detectors from academia and government perform relatively poorly, achieving a maximum AUC of only 74.78\% for images and 73.67\% for videos, suggesting that such detectors are \textbf{not yet suitable for direct public use} and require additional fine-tuning, development, and contexutalization to handle political deepfake scenarios effectively;} (3) \textbf{paid detectors generally outperform free-access counterparts}, potentially due to their continual updates, exposure to more diverse training data, and the integration of Retrieval-Augmented Generation modules (in LVLMs); (4) most LVLMs demonstrate stable performance and exhibit \textbf{minimal sensitivity} to common post-processing operations applied to the original media; and (5) \textbf{political video deepfake detection remains a persistent challenge}, with most detectors showing a marked drop in performance compared to image-level results. While careful threshold tuning can improve accuracy and reduce false acceptance rates, this approach demands expertise \textbf{beyond} that of average users. In general then, existing detectors \textbf{struggle to generalize} to authentic political deepfakes. 

Implications of these findings are significant. They bear on the suitability of both open and commercial detection tools as a solution for political misinformation, revealing gaps in the development and evaluation of these tools, as well as problematic disconnects if tool capabilities and limitations are poorly understood, including by platforms with core responsibilities to protect the public interest. Overall, detection of particular classes of inauthentic content constitutes important work, but if tools are not developed, translated, and implemented in a way that is responsive to the real-world context of political misinformation, they risk both failing to sufficiently address the core problem as well as providing false assurances when the other mitigation strategies be needed.

\begin{table*}[b!]
\centering
\small
\scalebox{0.8}{%
\begin{tabular}{ccccccccc}
\hline
\multirow{2}{*}{\textbf{Dataset}} & \multirow{2}{*}{\textbf{Year}} & \multicolumn{2}{c}{\textbf{Image}} & \multicolumn{2}{c}{\textbf{Video}} & \multirow{2}{*}{\textbf{\begin{tabular}[c]{@{}c@{}}Political \\  Deepfakes\end{tabular}}} & \multirow{2}{*}{\textbf{\begin{tabular}[c]{@{}c@{}}Human \\  Explanation\end{tabular}}} & \multirow{2}{*}{\textbf{\begin{tabular}[c]{@{}c@{}}In-the- \\  Wild\end{tabular}}} \\ \cline{3-4} \cline{5-6}
                                  &                                & \textbf{Real}        & \textbf{Fake}     & \textbf{Real}        & \textbf{Fake}      &                                                                         &                                                                                         \\ \hline
FaceForensics++ \cite{rossler2019faceforensics++}                   & 2019                           &       \xmark               &     \xmark      &\Checkmark &  \Checkmark       & \xmark                                                                        &   \xmark &\xmark                                                                                      \\
DeeperForensics-1.0 \cite{jiang2020deeperforensics}               & 2020                           &    \xmark                &  \xmark   &\Checkmark &     \Checkmark          &  \xmark                                                                       &   \xmark     &\xmark                                                                                 \\
Celeb-DF \cite{li2020celeb}                          & 2020                           &      \xmark            &   \xmark     &\Checkmark &   \Checkmark        &   \xmark                                                                      &  \xmark     &\xmark                                                                                          \\
DFDC \cite{dolhansky2020deepfake}                              & 2020                           &      \xmark            &  \xmark      &\Checkmark  &   \Checkmark        &   \xmark                                                                      &           \xmark     &\xmark                                                                                 \\
GenData \cite{teo2023measuring}                           & 2023                           & \xmark                       & \Checkmark           &\xmark & \xmark         &  \xmark                                                                       &            \xmark     &\xmark                                                                                \\
DF-Platter \cite{narayan2023df}                        & 2023                           &  \xmark                &  \xmark      & \Checkmark & \Checkmark        &      \xmark                                                                   &              \xmark     &\xmark                                                                              \\
DF40 \cite{yan2024df40}                              & 2024                           & \xmark                      & \Checkmark            &\xmark & \xmark    &  \xmark                                                                       &       \xmark     &\xmark                                                                                     \\

MMFakeBench \cite{liu2024mmfakebench}                       &  2024                              &   \xmark                    &    \Checkmark          &\xmark &  \xmark        &    \halfcheckmark                                                                     &             \xmark     &\xmark                                                                               \\ Deepfake-Eval \cite{chandra2025deepfake}                       &  2025                              &     \Checkmark                   &    \Checkmark          & \Checkmark  & \Checkmark             &  \halfcheckmark                                                                     &      \xmark     &\Checkmark                                                                                     \\AI-Face \cite{lin2025aiface}                           & 2025                           & \Checkmark                 & \Checkmark          & \xmark & \xmark         & \xmark                                                                       &      \xmark     &\xmark                                                                                     \\ \hline
\textbf{PDID (ours)}            & {2025}                           &\Checkmark     &\Checkmark   &\Checkmark                  &\Checkmark       & \Checkmark                                               &\Checkmark   & {\Checkmark }                                                        \\ \hline
\end{tabular}
}
\caption{\emph{Comparison of existing datasets with ours. Human explanation refers to the justification or rationale provided by a person to determine whether corresponding images or videos are real or fake.}}
\label{tab:datasets}
\end{table*}

\section{Results}\label{sec2}
\subsection{PDID Dataset Statistics and Analysis}
To demonstrate the significance and uniqueness of the PDID, we compare it against several widely used deepfake datasets.
(1) As shown in Table~\ref{tab:datasets}, the PDID is unique in its focus on politically salient deepfake `incidents' across both image and video modalities. In addition to manipulated media, it also includes authentic, real samples (e.g., images alleged to be deepfakes but confirmed to be authentic). 
(2) Unlike most existing datasets that either lack label explanations or rely on heuristically generated explanations (\eg, using LVLMs), our dataset includes human-verified labels with evidence of external verification by media or fact-checkers and textual explanations for both fake and real samples. This is a unique feature that enables future development of media forensics-specific LVLMs with reasoning using supervised fine-tuning. Although this application is beyond the scope of this work, it presents a promising direction for future research.
(3) Moreover, our dataset is constructed entirely from media disseminated in the wild—unlike most existing datasets, which rely on lab-generated content using known deepfake, GAN, or diffusion models. This makes our fake samples significantly more representative of the types of manipulations encountered in real-world scenarios, and more likely to have real-world impact. As such, the PDID dataset provides a more realistic and useful foundation for evaluating and designing deepfake detection methods in the political context (e.g., as opposed to the corporate fraud or intimate content context which may disseminate through different channels).

\begin{figure}[b]
    \centering
    \includegraphics[width=1\textwidth]{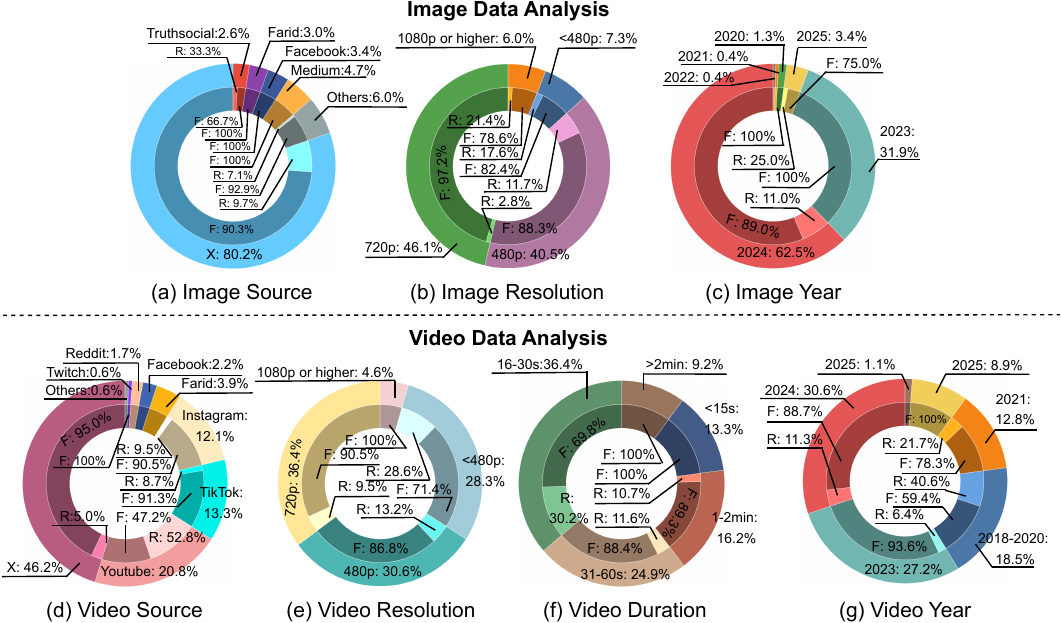}
    \caption{\emph{Dataset distribution across image and video sources, resolutions, durations, and years. The outer ring illustrates the overall proportion of each category, while the inner ring shows the breakdown of fake (F) and real (R) samples within each group.}} 
    \label{fig:dataset}
\end{figure}

\subsubsection{Dataset Distribution}
Since the PDID database is continually updated, for this benchmark study, we extract only the images and videos posted between 2018 and September 2025 that have undergone external verification and were subsequently labeled by both professional annotators and domain experts. This results in a curated dataset of 232 images and 173 videos. Additional details regarding the database and the extraction and labeling strategy are provided in Section~\ref{sec:PDID}. To better understand the characteristics of the extracted samples, we compute distribution statistics for both images and videos, with the results summarized in Figure~\ref{fig:dataset}. Specifically,
\begin{enumerate}
    \item \textbf{Image Data}. As shown in Figure~\ref{fig:dataset} top, most images originate from X (formerly Twitter) (Fig.~\ref{fig:dataset}(a)), while smaller portions are collected from Facebook, TruthSocial, Medium, a researcher-hosted website (Farid)~ \cite{Farid}, which archived deepfakes related to the 2024 U.S. Presidential Election, and other miscellaneous sources. Across all platforms, fake images constitute the majority of samples. In terms of resolution (Fig.~\ref{fig:dataset}(b)), most images are relatively low quality, with 720p (46.1\%) and 480p (40.5\%) being the most common. This may be due to post-processing applied by social media platforms, which often downscale uploaded images to improve loading speed and accessibility at the expense of pixel-level detail. Finally, the yearly distribution of fake images (Fig.~\ref{fig:dataset}(c)) shows that the majority were generated in 2023 and 2024, aligning with the 2024 U.S. Presidential Election, during which a surge of politically motivated deepfakes circulated on social media.

    \item \textbf{Video Data}. As shown in Figure~\ref{fig:dataset} bottom, the majority of videos (46.2\%) are sourced from X, with additional samples collected from YouTube, TikTok, and Instagram (Fig.~\ref{fig:dataset}(d)). Similar to the fake image distribution, fake videos represent the majority within most platform groups. In terms of resolution (Fig.~\ref{fig:dataset}(e)), low-resolution videos (below 720p) dominate, reflecting the post-processing and compression typically applied by social media platforms regardless of media modality. The duration distribution (Fig.~\ref{fig:dataset}(f)) further reveals that most videos (74.6\%) are shorter than one minute, with the majority of these being fake. This trend suggests that creators of deepfakes may prefer producing short videos, which are both more difficult for the public to scrutinize and technically easier to generate, given the current limitations of generative AI in producing long and highly realistic videos. Finally, Figure~\ref{fig:dataset}(g) shows the temporal distribution, which mirrors the trend observed in images: most deepfake videos were produced in 2023 and 2024. Notably, the proportion of deepfake videos increased from 18.5\% in 2018–2020 to 30.6\% in 2024. Although both 2020 and 2024 were U.S. election years, the sharp increase in 2024 can be attributed to the wider availability and accessibility of generative AI tools (\eg, Sora 2 \cite{sora2}), making deepfake creation significantly easier and more prevalent.
\end{enumerate}

\subsubsection{Power Spectrum Analysis}
These diverse characteristics reflect content distributions encountered in real-world settings, which differ significantly from those found in lab-based synthetic datasets. To better understand the substantial differences between real-world fake images (from the PDID) and lab-generated fake images, we conduct a power spectrum analysis—a technique that reveals signal patterns in the frequency domain while minimizing the influence of semantic content. This analysis compares images from the PDID, conventional deepfake video datasets, GAN-generated datasets, and DM-generated datasets. The latter three categories are sourced from the AI-Face dataset~ \cite{lin2025aiface}. Real images corresponding to each category for the fake image generation are also included for comparison.  
Additional details are provided in Section~\ref{sec:frequecy}.


\begin{figure}[t]
    \centering
        \includegraphics[width=1\linewidth]{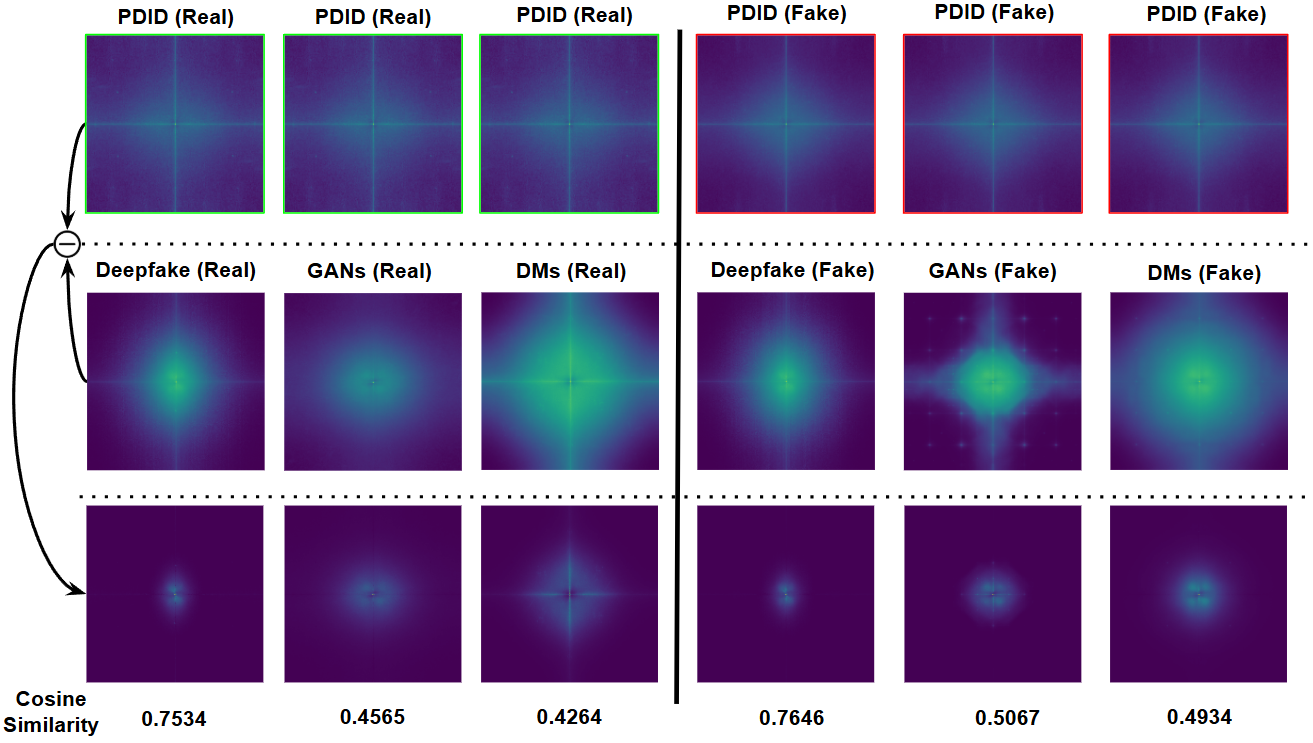}
        \caption{\emph{(Left three columns) Comparison of power spectra between the PDID real and others real. (Right three columns) Comparison of power spectra between the PDID fakes and other deepfakes. The bottom figures show the differences between the corresponding spectra in the top two rows, with the cosine similarity quantifying similarity.}}
    \label{fig:real_comparison}
\end{figure}

\textbf{PDID (Real) vs Others (Real)}.
Figure~\ref{fig:real_comparison} (Left) compares the spectra of real images from the PDID with those from the other three categories. We conduct both qualitative and quantitative (\ie, cosine similarity, where higher values indicate stronger alignment in spectral patterns) comparisons. Both visual analysis and cosine similarity values reveal a clear distinction: the real samples in the PDID differ substantially from those in the Deepfake, GAN, and DM categories. This divergence is likely due to the constrained nature of the real data used in the latter categories:
\begin{enumerate}
    \item In the Deepfake category, real videos are typically sourced from YouTube clips featuring trackable, mostly frontal faces with minimal occlusion, or recorded under controlled conditions with consented individuals. Some focus on celebrity interviews. These constraints limit diversity and reduce the authenticity typically observed in real-world social media content.
    \item In the DM category, real images used to train diffusion models are selected from the IMDB-WIKI dataset~ \cite{rothe2015dex}, which primarily consists of celebrity portraits. Moreover, this dataset was released in 2015 and does not reflect the current visual characteristics or resolutions commonly found in modern social media images.
    \item In the GAN category, real images used to train GANs come from the FFHQ dataset~ \cite{karras2019style}, which contains high-resolution (1024×1024), high-quality PNG images with diverse age, ethnicity, and background variations. However, this stands in contrast to the real images in the PDID, which are often heavily compressed and typically fall below 720p resolution, as shown in the resolution distributions (see Figure \ref{fig:dataset}). These compression artifacts and resolution differences contribute significantly to the divergence in power spectral characteristics.
\end{enumerate}
Interestingly, among the three categories, the Deepfake real images appear most similar to those in the PDID, as evidenced by its relatively higher cosine similarity score. This may be because many of the real videos in the PDID are also sourced from YouTube, aligning more closely with the data origin of the Deepfake category compared to the other two.

\textbf{PDID (Fake) vs Others (Fake)}.
Figure~\ref{fig:real_comparison} (Right) compares the spectra of fake images in the PDID with those from the other three categories. Both visual analysis and cosine similarity scores reveal substantial differences, indicating that the PDID fake content exhibits distinct spectral characteristics. Several factors may explain this divergence:
\begin{enumerate}
    \item Fake images and videos in the PDID are sourced from real-world social media platforms and may be produced by a wide range of generative AI tools. Many of them may not be represented in the training or generation pipelines of the other three categories. As such, the generative characteristics of the PDID fakes are more heterogeneous.\footnote{Notably, we witness this heterogeneity in the PDID dataset despite some areas of uniformity, as noted in the limitations section. For instance, PDID images and videos are 1) largely sourced from the US rather than a global context, 2) focus on a modest subset of contexts (\eg, political events or figures rather than entertainment or sports), and 3) feature prominent individuals like political party leaders in a plurality of cases. It is striking that the PDID is potentially more structurally diverse than traditional laboratory datasets used for deepfake detection despite these built-in semantic constraints.}
    \item Unlike lab-generated media, which typically undergo little to no post-processing, political fake media in the PDID from social media often experiences multiple stages of transformation before dissemination. These could include compression, editing, filtering, and platform-specific re-encoding, all of which introduce additional artifacts that alter the frequency-domain properties of the media.
\end{enumerate}
These observations demonstrate the unique challenges posed by real-world settings, as captured in the PDID dataset. Traditional detectors primarily developed and evaluated on lab-generated datasets may struggle to detect real-world political deepfakes. This motivates our systematic evaluation of existing detection models on the PDID dataset to expose current limitations and identify areas for improvement. These challenges and findings are further explored in the following sections.

\subsection{Benchmark Evaluations}
In this section, we evaluate existing, state-of-the-art deepfake detectors provided by academic researchers, government initiatives, and commercial providers. While several prior studies have conducted benchmarking, our work differs significantly in scope. Specifically, we focus on the detection of real-world political deepfakes. Moreover, existing benchmarks typically focus on a narrow range of (academic) models, whereas we expand coverage by including detectors from government agencies and industry providers, which have been largely overlooked in past efforts (see Table~\ref{tab:benchmark-comparison}). By incorporating this broader set of detectors, our evaluation exposes both practically significant limitations that emerge when detectors are deployed in real-world settings and category-specific weaknesses that can inform directions for future improvement.

To better differentiate detector types, we organize them into two high-level categories based on whether large vision language models (LVLMs) are incorporated in their development: LVLM-agnostic detectors and LVLM-aware detectors. The LVLM-agnostic group is further divided into white-box detectors and black-box detectors (\ie, commercial tools), depending on the availability of source code and implementation details. Within the white-box group, we additionally distinguish between detectors originating from academia and those specifically facilitated by government programs.

\begin{table}[t]
\centering
\scalebox{0.87}{%
\begin{tabular}{ccccccc}
\hline
\multirow{3}{*}{\textbf{Benchmark}} & \multirow{3}{*}{\textbf{Year}}& \multirow{3}{*}{\textbf{\begin{tabular}[c]{@{}c@{}}Real-world\\Political \\Deepfakes\end{tabular}}} & \multicolumn{3}{c}{\textbf{LVLM-Agnostic Detectors}}         & \multirow{3}{*}{\textbf{\begin{tabular}[c]{@{}c@{}}LVLM-Aware \\ Detectors\end{tabular}}} \\ \cline{4-6}
                                    &         &                       & \multicolumn{2}{c}{\textbf{White Box}}  & \textbf{Black Box}  &                                                                                          \\
                      &              &                                & \textbf{Academia} & \textbf{Government} & \textbf{Commercial} &                                                                                          \\ \hline
Loc et al. \cite{trinh2021examination}                          & 2021    &           \xmark            & \checkmark         &       \xmark              &          \xmark             &   \xmark                                                                                         \\
DeepfakeBench \cite{DeepfakeBench_YAN_NEURIPS2023}                       & 2023       &         \xmark           & \checkmark         & \xmark                      & \xmark                      &    \xmark                                                                                        \\
CDDB \cite{li2023continual}                                & 2024             &   \xmark           & \checkmark         &       \xmark                &    \xmark                   &                   \xmark                                                                         \\
Deng et al. \cite{deng2024towards}                         & 2024       &        \xmark            & \checkmark         &             \xmark          &  \xmark                     &    \xmark                                                                                        \\
DF40 \cite{yan2024df40}                                & 2024         &           \xmark       & \checkmark         &  \xmark                     &     \xmark                  &      \xmark                                                                                      \\
MMFakeBench \cite{liu2024mmfakebench}                         & 2024      &         \halfcheckmark             &   \xmark        &      \xmark                 &    \xmark                   & \checkmark                                                                                \\
AI-Face \cite{lin2025aiface}                             & 2025             &  \xmark             & \checkmark         &     \xmark                  &      \xmark                 &     \xmark                                                                                       \\ 
Shield \cite{shi2025shield}                             & 2025             &     \xmark           &    \xmark        &\xmark                       &   \xmark                    &       \checkmark                                                                                   \\ 
Ren et al. \cite{ren2025can}                             & 2025            &     \xmark            &   \xmark         &        \xmark               &    \xmark                   &       \checkmark                                                                                   \\\hline
\textbf{Ours}                & 2025        &   \checkmark                & \checkmark         & \checkmark           & \checkmark           & \checkmark                                                                                \\ \hline
\end{tabular}
}
\caption{\emph{Comparison of existing deepfake detection benchmarks and ours.}}
\label{tab:benchmark-comparison}
\end{table}

\subsubsection{LVLM-Agnostic White-Box Detectors}
\begin{table}[t]
\centering
\scalebox{0.9}{%
\begin{tabular}{c|c|c|ccc|ccc}
\hline
\multirow{2}{*}{\begin{tabular}[c]{@{}c@{}}Detector \\ Type\end{tabular}} & \multirow{2}{*}{Category}                                                     & \multirow{2}{*}{\begin{tabular}[c]{@{}c@{}}Detector \\ Name\end{tabular}} & \multicolumn{3}{c|}{Image}                        & \multicolumn{3}{c}{Video}                         \\
                                                                          &                                                                               &                                                                           & ACC$\uparrow$  & AUC$\uparrow$  & FAR$\downarrow$ & ACC$\uparrow$  & AUC$\uparrow$  & FAR$\downarrow$ \\ \hline
\multirow{12}{*}{Academia}                                                & \multirow{3}{*}{Naive}                                                        & Xception~\cite{chollet2017xception}                 & 84.57          & 64.30          & 8.67            & 80.63          & 60.09          & 3.03            \\
                                                                          &                                                                               & EfficientNet-B4~\cite{tan2019efficientnet}          & 81.38          & 67.09          & 16.18           & 81.25          & 72.17          & 4.55            \\
                                                                          &                                                                               & ViT-B/16~\cite{vit}                                 & 87.77          & 66.36          & 5.78            & 81.25          & 67.38          & \textbf{1.52}   \\ \cline{2-9} 
                                                                          & \multirow{3}{*}{Frequency}                                                    & F3Net~\cite{qian2020thinking}                      & \textbf{89.36} & 71.54          & \textbf{2.89}   & 80.00          & 57.41          & 5.30            \\
                                                                          &                                                                               & SPSL~\cite{liu2021spatial}                         & 88.83          & 63.47          & 6.36            & 78.75          & 58.89          & 6.06            \\
                                                                          &                                                                               & SRM~\cite{luo2021generalizing}                      & 78.72          & \textbf{74.78} & 18.50           & \textbf{82.50} & 69.82          & 3.79            \\ \cline{2-9} 
                                                                          & \multirow{3}{*}{Spatial}                                                      & UCF~\cite{yan2023ucf}                              & 86.70          & 72.00          & 6.36            & 78.12          & 60.13          & 8.33            \\
                                                                          &                                                                               & UnivFD~\cite{ojha2023towards}                      & 52.66          & 48.82          & 46.24           & \textbf{82.50} & 69.81          & 4.55            \\
                                                                          &                                                                               & CORE~\cite{ni2022core}                              & 52.66          & 57.96          & 46.24           & 80.00          & 66.52          & 8.33            \\ \cline{2-9} 
                                                                          & \multirow{3}{*}{\begin{tabular}[c]{@{}c@{}}Fairness-\\ enhanced\end{tabular}} & DAW-FDD~\cite{ju2024improving}                      & 64.36          & 66.15          & 35.26           & \textbf{82.50} & 68.51          & 3.79            \\
                                                                          &                                                                               & DAG-FDD~\cite{ju2024improving}                     & 85.64          & 52.60          & 6.94            & 81.87          & 54.73          & \textbf{1.52}   \\
                                                                          &                                                                               & PG-FDD~\cite{lin2024preserving}                    & 85.11          & 65.47          & 8.67            & 80.63          & \textbf{73.67} & 4.55            \\ \hline
\multirow{3}{*}{Government}                                               & \multirow{3}{*}{-}                                                            & CNNDetection~\cite{wang2020cnn}                     & 22.87          & 42.52          & 81.50           & 16.27          & 65.95          & 97.20           \\
                                                                          &                                                                               & KitwareDetector~\cite{KitwareDetector}               & 8.51           & 57.23          & 99.42           & 13.86          & 53.50          & 100.00          \\
                                                                          &                                                                               & GANattribution~\cite{ganattribution}                 & 22.87          & 29.75          & 82.08           & 67.47          & 50.55          & 25.87           \\ \hline
\end{tabular}
}
\caption{\emph{Performance (\%) comparison for LVLM-agnostic white-box detectors. The best results are shown in \textbf{Bold}.}}
\label{tab:academia_and_government}
\end{table}
We first evaluate twelve widely used academic deepfake detectors spanning four major categories (\ie, naive backbones, frequency-based, spatial-based, and fairness-enhanced), all pretrained on the AI-Face dataset~\cite{lin2025aiface}. In addition, we assess three detectors developed under the DARPA Semantic Forensics (SemaFor) program~\cite{darpa_semafor}. The performance of these models is summarized in Table~\ref{tab:academia_and_government}. (1) Overall, all detectors exhibit low Accuracy (ACC) and Area Under the Receiver Operating Characteristic Curve (AUC) in both image and video settings. For instance, the highest AUC achieved is only 74.78\% for images and 73.67\% for videos. These results indicate that \textbf{detectors developed in academia and government are not sufficiently effective for detecting real-world political deepfakes disseminated on social media}. (2) When comparing detectors developed by academia and government, it is evident that academic detectors achieve substantially higher ACC and AUC scores than government-developed detectors. One possible explanation is that government detectors were trained on datasets composed of images and videos generated by a limited set of generative models under controlled conditions, typically restricted to a single generative architecture curated within the program. In contrast, the academic detectors evaluated here were trained on a more recent dataset that includes samples produced by a wide range of generative models. Additionally, since the DARPA SemaFor program concluded some time ago, the corresponding detectors may no longer receive updates or maintenance, further contributing to their relatively lower performance on contemporary, real-world political deepfakes. This finding urges that \textbf{continuous updates and adaptation of deepfake detection models are indispensable}, as detectors developed only a few years ago appear to be largely obsolete when confronted with evolving generative techniques.
(3) Among the academic detectors, F3Net~\cite{qian2020thinking} and SPSL~\cite{luo2021generalizing} achieve the strongest relative performance at the image level. A plausible explanation is that both methods operate in the frequency domain, making them \textbf{less sensitive to semantic content} (\eg, particular individual politicians depicted in an image) and more focused on identifying \textit{signal-level artifacts}. Consequently, the actual deepfake content contributes less to detection performance than underlying artifacts or traces produced by the generative model; these dominate the detector’s decision process.  
Interestingly, all academic detectors demonstrate comparable ACC and favorable False Acceptance Rate (FAR) performance on the video dataset. As detailed in Section \ref{sec:benchmark-settings}, we randomly extract 10 frames from each video for testing and develop an aggregation approach to determine the final video prediction: temporal consistency or averaging effects across frames may help stabilize predictions and explain the relatively strong predictive performance here.

\begin{figure}[t]
    \centering
    \includegraphics[width=1\textwidth]{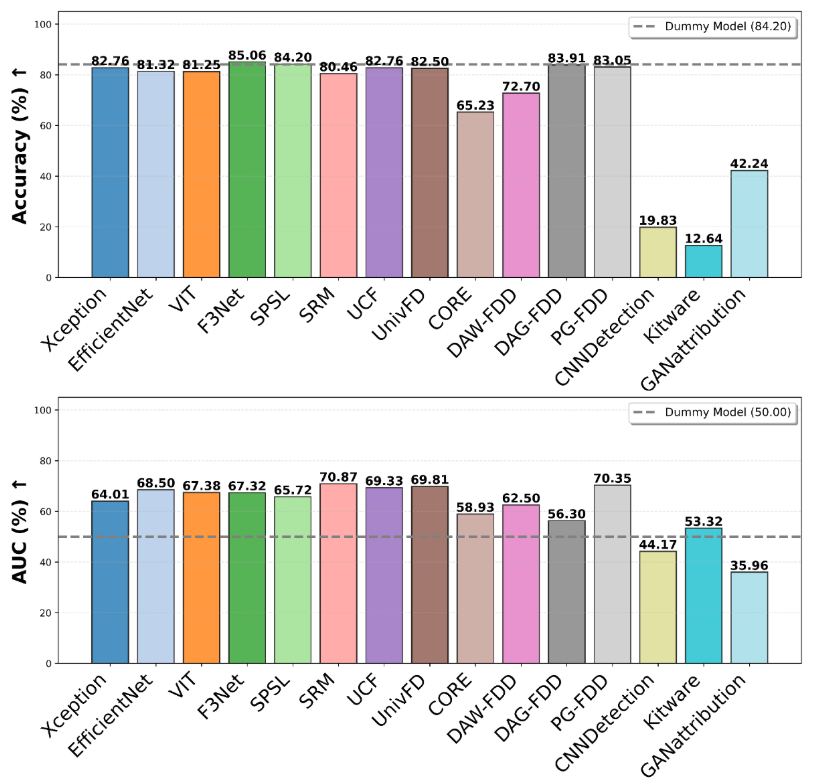}
    \caption{\emph{Performance comparison between LVLM-agnostic white-box detectors and “dummy model” on the image and video mixed set. The “dummy model” refers to a baseline that predicts all samples as fake. } }
    \label{fig:academia_government}
\end{figure}

We next combine the image and video sets into a single dataset and evaluate all detectors on this unified collection. To better understand their performance, we introduce a dummy baseline model (always fake prediction) that predicts every sample as fake with 100\% probability, and compare all detectors against it. The dummy baseline represents a simple yet revealing reference point: it predicts every sample as “fake.” While seemingly trivial, it captures an important real-world phenomenon—if a large portion of the media circulating online or within certain domains (such as political misinformation) is indeed fake, then a naive system that always flags content as fake may appear to perform competitively under conventional metrics. This highlights a key limitation of current evaluation practices: detectors optimized for high sensitivity (flagging many fakes) can achieve deceptively strong results without true discernment. In practice, such imbalance could create a perverse incentive for developers to tune models toward over-detection, masking poor specificity, falsely assuring users, and undermining public trust in automated detection systems. The results are summarized in Figure~\ref{fig:academia_government}. Only one detector (\ie, F3Net \cite{qian2020thinking}) surpasses the dummy baseline in terms of ACC. While most detectors achieve higher AUC scores than the dummy model, notable exceptions emerge among the government-developed detectors. Specifically, CNNDetection~\cite{wang2020cnn} and GANattribution~\cite{ganattribution}, which tend to assign higher confidence scores to real samples than to fake ones, effectively predict in the wrong direction (worse than a random guesser).

\begin{figure}[t]
    \centering
    \includegraphics[width=1\linewidth]{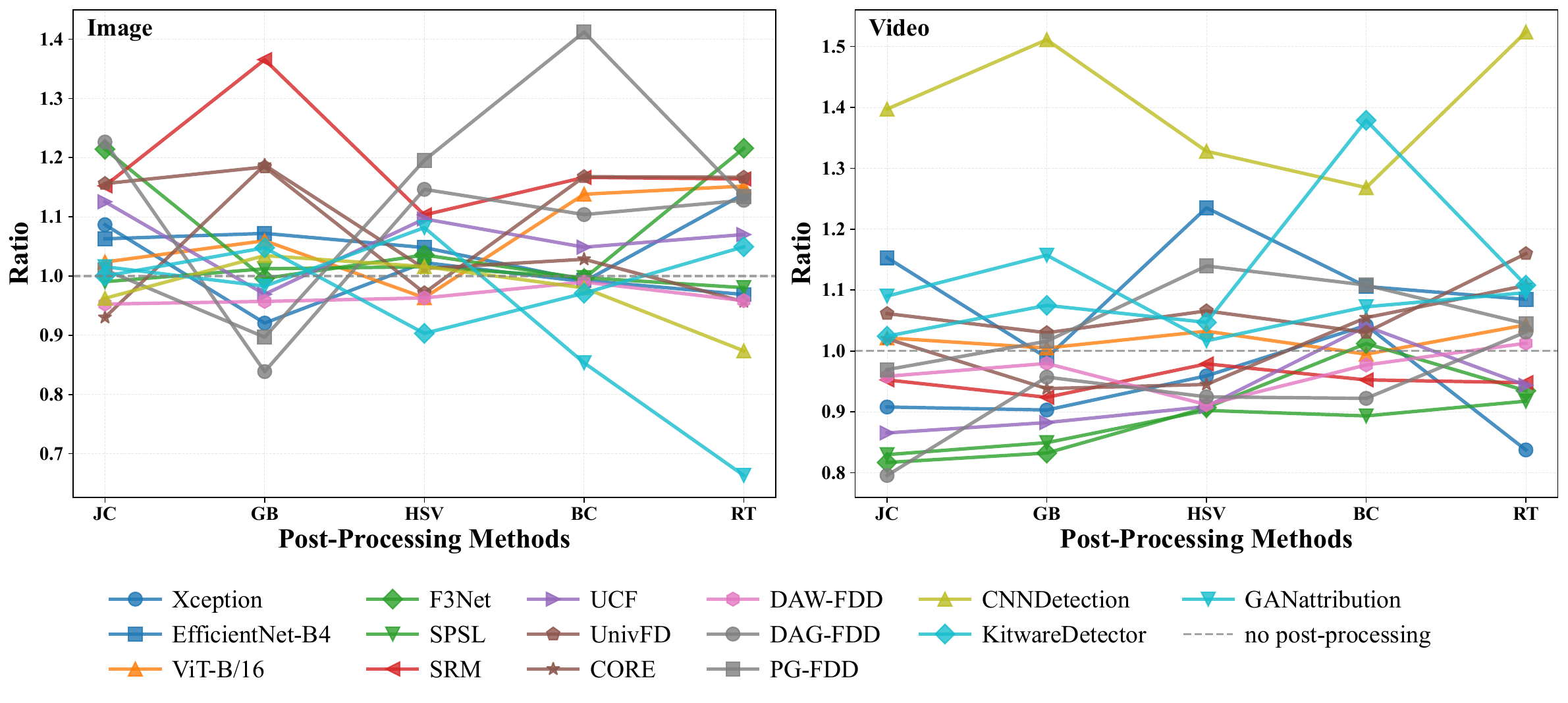}\\[1em] 
    \caption{\emph{Performance (AUC) ratio (original vs. post-processed) for LVLM-agnostic white-box detectors.}}
    \label{fig:white-robustness_test}
\end{figure}

However, \textbf{the low performance of these models does not necessarily imply that they are poorly designed; rather, it reflects their limited suitability for detecting real-world political deepfakes}. Many of these detectors were originally developed to target specific manipulation types, such as CNN- or GAN-generated media, under controlled conditions. Their underperformance is thus expected when faced with the unpredictable, mixed, and evolving manipulations found in social media. In contrast to the laboratory setting, users in the real world are not informed whether an image or video is GAN-generated, diffusion-based, or authentic, which exposes a critical mismatch between benchmark-specific model capabilities and practical utility. This discrepancy reiterates the importance of evaluating detectors from the perspective of ordinary users, who rely on them for trustworthy assessments without technical context. Compounding this issue, some commercial tools (as discussed further in the next section) may overstate their effectiveness while downplaying their limitations\footnote{For example, one company advertises localization capabilities that remain non-functional even in upgraded plans, while several companies claim more than 98 or 99\% accuracy, whereas four of the tools we test are below 65\% accuracy.}.

We also conduct a robustness assessment of all detectors by applying five post-processing operations (\ie, JPEG Compression (JC), Gaussian Blur (GB), HSV Shift (HSV), Brightness/Contrast Adjustment (BC), Rotation (RT)) as detailed in Section~\ref{sec:benchmark-settings}. The results are presented in Figure~\ref{fig:white-robustness_test}. For both image and video sets, we find that most performance change ratios (with post-processing versus without) fall within the range of [0.9, 1.1], indicating that \textbf{post-processing has minimal impact on the detectors’ effectiveness}. This stability may be attributed to the fact that political deepfake samples in the PDID are sourced from social media, where they have already undergone various post-processing operations such as compression and blurring. Consequently, applying additional post-processing introduces little further change in detector performance. However, we indeed find a few detectors that are sensitive to specific post-processing approaches.


\subsubsection{LVLM-Agnostic Black-Box Detectors}
Recognizing that LVLM-agnostic white-box detectors from academia and government may have inherent limitations (such as training data constraints and lack of ongoing maintenance), our findings could fail to capture state-of-the-art practice in deepfake detection. To address this, we evaluate several of the most widely used commercial deepfake detection tools. Although these tools are not open-source, they are typically actively maintained and regularly updated, making them more representative of current real-world detection capabilities, particularly for ostensibly publicly-important (or commercializable) domains like deepfakes used for commercial fraud or political misinformation. To ensure a fair comparison, we conducted all tests within a short time frame (September 2025) to minimize the impact of version updates and promote consistency across evaluations.

Table~\ref{tab:commercial_tool_results} presents a comparative evaluation of LVLM-agnostic black-box detectors (\ie, commercial tools) across both image and video modalities. The comparison includes widely used platforms such as \textit{Incode}, \textit{Reality Defender}, \textit{Hive Moderation}, and others. These commercial detectors operate as closed-source platforms, each with distinct input requirements, rejection criteria, and output formats, which collectively affect their coverage, usability, and interpretability. For instance, \textit{Incode} is a face-oriented deepfake detector primarily designed for identity verification and biometric authentication. It automatically rejects inputs when no face is detected or when the detected region is too small, resulting in partial sample exclusion. Similarly, \textit{BrandWell} accepts only images in .jpg, .png, or .webp formats and discards all others, reducing its effective sample size for our evaluation. However, these constraints should not affect the overall conclusions of our evaluation, as we assess the tools from the perspective of average users—those without technical expertise who typically rely on such platforms and predictions at face value, and are not positioned to perform extensive testing, tweaking, or interpretation across multiple API endpoints, and so on. 

\begin{table}[t]
\scalebox{0.92}{
\begin{tabular}{c|c|cccccc}
\hline
\multirow{2}{*}{\makecell{Data \\  Modality}} & \multirow{2}{*}{\makecell{Commercial\\ Tools}} & \multirow{2}{*}{ACC $\uparrow$} & \multirow{2}{*}{AUC $\uparrow$} & \multirow{2}{*}{FAR $\downarrow$} & \multicolumn{2}{c}{Output}    & \multirow{2}{*}{Date} \\ 
                  &                   &                   &                   &                   & \multicolumn{1}{c}{Binary} & Probability &                   \\ \hline
\multirow{9}{*}{Image} &          Incode*~\cite{incode} & 91.07 & 52.70 & \textbf{2.56} & \checkmark & \checkmark & 9/13/2025                  \\ 
                  &            Is It AI~\cite{isitai} & 89.61 & \textbf{96.15} & 12.00 & \checkmark & \checkmark & 9/14/2025                  \\ 
                  &             Winston~\cite{gowinston} & 61.64 & 85.73 & 43.50 & \xmark & \checkmark & 9/14/2025 \\  
                  &            BrandWell*~\cite{brandwell_ai_image_detector} & 59.02 & 39.34 & 31.58 & \xmark & \checkmark & 9/14/2025 \\  
                  &             AI or Not~\cite{aiornot} & 89.22 & 95.01 & 12.00 & \checkmark & \checkmark & 9/15/2025 \\ 
                  &                 Illuminarty~\cite{illuminarty_ai} & 49.37& 70.48 & 57.07 & \xmark & \checkmark & 9/15/2025 \\ 
                  &                Reality Defender~\cite{realitydefender} & \textbf{93.51} & 95.82 & 5.53 & \checkmark & \checkmark & 9/15/2025 \\  
                  &           Hive Moderation~\cite{hivemoderation} & 83.62 & 94.47 & 18.50 & \checkmark & \checkmark & 9/16/2025 \\
                  &                Deepfake Detector~\cite{deepfakedetector} & 49.57 & 68.59 & 58.50 & \checkmark & \checkmark & 9/16/2025                  \\ \hline
\multirow{4}{*}{Video} &          Incode*~\cite{incode} &\textbf{77.27}  &44.74  &\textbf{10.53}  &\checkmark  &\checkmark  &   9/24/2025               \\ 
                  &     AI or Not~\cite{aiornot} & 75.14 & 59.42 & 17.02 & \checkmark & \checkmark & 9/18/2025                 \\ 
                  &    Hive Moderation~\cite{hivemoderation} & 44.51 & \textbf{77.98} & 65.96 & \xmark & \checkmark & 9/18/2025                  \\ \hline
\end{tabular}
}
\caption{\emph{Performance (\%) comparison for LVLM-agnostic black-box detectors (\ie, commercial tools). * indicates that the corresponding tool rejects some test samples.}}
\label{tab:commercial_tool_results}
\end{table}

Understanding their performance on political deepfakes is therefore particularly valuable, representing an instance where regular users (perhaps mediated through social media platforms) are relying on a detection tool fairly straightforwardly, as opposed to when commercial tools are used by corporate fraud experts, journalists, or forensic experts, for example, who have the expertise and resources to probe deeply into the suitability and interpretation of various detection tools. \textbf{Finally, it is important to note that our goal is not to determine which tool performs `best' in an abstract sense, but rather to highlight the challenges and limitations these commercial systems face when applied to the detection of real-world political deepfakes}. This caveat is critical, as the tested dataset, determination of labels for authentic and inauthentic content, and particular evaluation regimes are at least partially subjective. For instance, even the determination of whether a given video is authentic or inauthentic is complicated (\eg, a real newscast presenting coverage of a video deepfake, or an authentic image with a deceptive text caption). As a result, \textbf{we encourage readers to interpret results holistically, not limiting their attention to ranking the `best' tools. Such comparisons require a dynamic and contextual understanding not captured by the quantitative metrics here alone}.

\begin{figure}[t]
    \centering
    \includegraphics[width=1\textwidth]{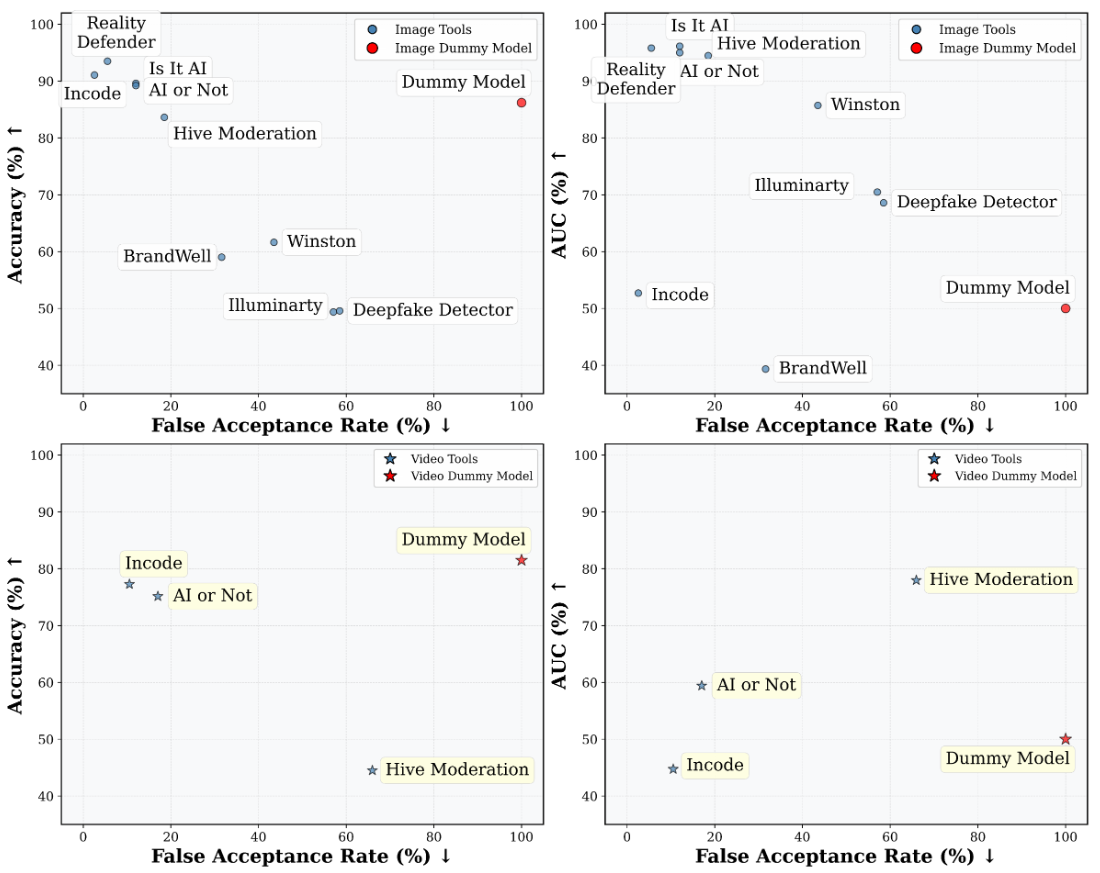}
    \caption{\emph{Performance comparison for commercial tools on the image set (Top) and video set (Bottom). The “dummy model” refers to a baseline that predicts all samples as fake. }} 
    \label{fig:commercial-tool}
\end{figure}

As shown in Table~\ref{tab:commercial_tool_results}, at the image level, although only two tools achieve an accuracy above 90\%, four tools obtain AUC values exceeding 90\%. However, most tools perform less effectively in terms of FAR, with values above 10\%. Specifically, \textit{Reality Defender} achieves the highest overall accuracy (93.51\%) and a strong AUC of 95.82\%, indicating excellent discriminative ability in detecting political deepfakes. \textit{Is It AI} and \textit{AI or Not} also perform competitively, with AUCs of 96.15\% and 95.01\%, respectively, demonstrating reliable generalization. \textit{Incode} attains a high accuracy of 91.07\% while maintaining the lowest FAR (2.56\%), reflecting a well-calibrated decision boundary that minimizes false positives, a particular concern and priority given the importance of discernment between true and false information in political contexts. In contrast, \textit{BrandWell} and \textit{Deepfake Detector} exhibit notably weaker performance, with AUCs of 39.34\% and 68.59\%, respectively, suggesting limited robustness and poor generalization when applied to political deepfake detection, at least in the context of this study's evaluation approach and dataset.

At the video level, all three tools demonstrate lower performance compared to their results on the image set. \textbf{This indicates that detecting deepfakes in videos remains more challenging than in static images for existing commercial tools}. Among the three tools, \textit{Incode} achieves the highest accuracy (77.27\%), followed closely by \textit{AI or Not} (75.14\%), both maintaining a reasonable balance between detection reliability and false alarm rates. \textit{Hive Moderation} attains the highest video-level AUC (77.98\%); however, its elevated FAR (65.96\%) suggests a strong tendency to misclassify real videos as fake, limiting its practical reliability in political video deepfake detection.

We also include dummy baseline models for both images and videos (100\% of content identified as fake), as done previously for the LVLM-agnostic white-box detectors, to better understand the relative performance of the commercial tools. As shown in Figure~\ref{fig:commercial-tool}, only four tools outperform the dummy model in terms of ACC on the image set. However, all tools perform worse than the dummy model in terms of ACC on the video set, highlighting the additional difficulty of video-based deepfake detection. In contrast, most tools achieve substantially higher AUC values than the dummy model, indicating that while their decision thresholds may not be optimal, they still retain some discriminative capability.

A major epistemic and practical problem remains, however: while the latent capability of these models may exceed their measured accuracy at a standard threshold (\eg, probability is greater than 0.5 for a binary determination), users and platforms are rarely in a position to meaningfully or dynamically calibrate. For instance, users of detection tools often lack access to reliable ground truth by definition, making proactive calibration difficult—if not impossible—in real-world settings. Moreover, the distribution of deepfake types continually shifts as new generation techniques, compression artifacts, and media modalities emerge, rendering optimized thresholds potentially brittle over time. Finally, contextual stakes complicate decision-making: it is unclear what the acceptable balance between false positives and false negatives should be in contexts as varied as social media moderation across diverse platforms, forensic analysis, or journalistic verification. In short, \textbf{calibration presupposes epistemic stability—yet neither the data environment nor common institutional settings provides a clear basis for knowing where, or how, to draw the line}. As these uncertainties play out in practice, platforms risk (unknowingly or untestably) deploying oversensitive or miscalibrated systems that either overflag authentic content or miss harmful manipulations, producing uneven performance across contexts and unavoidably exposing the public to harms.



\begin{table}[t]
\centering
\scalebox{1}{%
\begin{tabular}{c|c|ccc|ccc}
\hline
\multirow{2}{*}{Cost} & 
\multirow{2}{*}{LVLM} & 
\multicolumn{3}{c|}{Images} & 
\multicolumn{3}{c}{Videos} \\ 
\hhline{~~------}
& & ACC$\uparrow$ & AUC$\uparrow$ & FAR$\downarrow$ & ACC$\uparrow$ & AUC$\uparrow$ & FAR$\downarrow$ \\ 
\hline
& ChatGPT(GPT-5) \cite{openai2025gpt5} & 84.55 & \textbf{96.63} & 18.00 & 68.79 & \textbf{85.28} & 36.88 \\
\multirow{2}{*}{Paid} 
& Qwen-VL-Max* \cite{Qwen-VL} & 81.45 & 89.77 & 20.11 & 68.79 & 71.79 & 32.62 \\ 
& Claude-Sonnet-4.5* \cite{anthropic2025claude45} & 87.50 & 93.01 & 14.00 & 73.99 & 80.04 & 27.66 \\
& Gemini-2.5-pro* \cite{comanici2025gemini} & \textbf{88.79} & 65.48 & 6.50 & 73.99 & 79.70 & 24.82 \\
\hhline{-|-------}
& LLaVA-v1.5-7B \cite{lin2023video} & 86.21 & 54.31 & \textbf{0.00} & 68.79 & 31.33 & 16.31 \\
& LLaVA-v1.5-13B \cite{liu2023llava} & 86.21 & 56.34 & \textbf{0.00} & \textbf{81.50} & 27.07& \textbf{0.00} \\
\multirow{2}{*}{Free} 
& LLaVA-v1.5-13B-XTuner \cite{xtuner2023} & 39.11 & 55.91 & 67.88 & 79.77 & 38.61 &2.13 \\
& DeepSeek-VL-7B \cite{lu2024deepseek} & 78.45& 67.86 & 16.00 & 69.36 & 51.77 & 23.40 \\
& CogVLM-Chat \cite{wang2024cogvlm} & 84.89 & 74.36 & 11.28 & 78.03 & 49.08 &4.96 \\ 
& Monkey-Chat \cite{Li_2024_CVPR} & 40.00 & 57.36 & 66.15 & 65.32 & 43.66 & 22.70 \\ 
\hline
\end{tabular}
}
\caption{\emph{Performance (\%) comparison for LVLM-aware detectors.* indicates that the corresponding detector not generate correct format response for some test samples.}}
\label{tab:vlm_results}
\end{table}

\subsubsection{LVLM-Aware Detectors}
Recently, Large Vision-Language Models (LVLMs) have gained popularity by enabling users to upload visual media and query them directly through natural language. Many paid LVLMs regularly updated and fine-tuned with new and diverse data, improving their adaptability and alignment with recent real-world content—similar to the continuous maintenance in commercial detection tools. Beyond this, they also incorporate a Retrieval-Augmented Generation (RAG) module \cite{gao2023retrieval}, which allows them to access and analyze up-to-date online information and integrate the retrieved evidence into their reasoning process, potentially producing more contextually grounded and accurate responses. Motivated by this capability, we evaluate the effectiveness of current LVLMs in detecting political deepfakes. In this experiment, we assess ten widely used LVLMs—four requiring paid API access and six freely available online. 

\begin{figure}[t]
    \centering
    \includegraphics[width=1\textwidth]{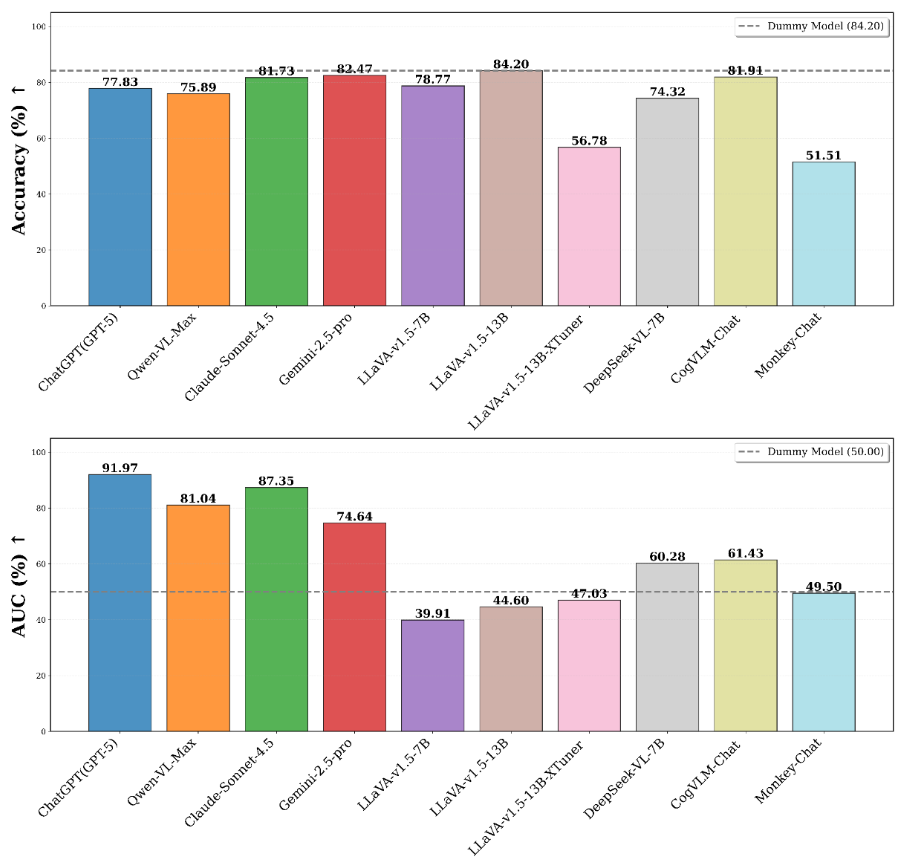}
    \caption{\emph{Performance comparison between LVLM-aware detectors and “dummy model” on the image and video mixed set. The “dummy model” refers to a baseline that predicts all samples as fake.}} 
    \label{fig:lvlm_accuracy_auc_far}
\end{figure}

The performance results for both image and video evaluations are summarized in Table~\ref{tab:vlm_results}.
(1) In general, none of the evaluated LVLMs achieve over 90\% accuracy across both data modalities. However, in terms of AUC, two LVLMs exceed 90\% performance on the image set. Interestingly, LLaVA-v1.5-7B~\cite{lin2023video} and LLaVA-v1.5-13B~\cite{liu2023llava} achieve a 0\% False Acceptance Rate (FAR) on images, with LLaVA-v1.5-13B maintaining this 0\% FAR on videos as well. This outcome suggests that LLaVA-v1.5-13B classifies all samples as fake, effectively yielding no false acceptances but at the cost of high false rejections.
(2) We also observe that \textbf{the paid LVLMs generally achieve higher AUC performance compared to the free-access models}. In contrast, most free LVLMs perform close to random guessing, with AUC values hovering around 50\%, indicating limited capability in distinguishing real from fake political media.
(3) Among all evaluated LVLMs, ChatGPT (GPT-5) demonstrates the best overall balance, achieving a strong image-level AUC of 96.63\% and video-level AUC of 85.28\%, indicating reliable discriminative performance across both modalities. Claude-Sonnet-4.5 follows closely, with competitive AUCs of 93.01\% for images and 80.04\% for videos, suggesting good generalization capability. However, both LVLMs exhibit relatively high False Acceptance Rates (FARs), which could mislead general users and increase the risk of misjudging political deepfakes as authentic.

We also combine the image and video sets into a single dataset and introduce a dummy baseline model (an ``always-fake'' predictor) to better understand the overall performance of LVLMs in detecting real-world political deepfakes. The results, presented in Figure~\ref{fig:lvlm_accuracy_auc_far}, show that none of the LVLMs outperform the dummy model in terms of Accuracy. As discussed earlier, LLaVA-v1.5-13B~\cite{liu2023llava} predicts all samples as fake, thereby matching the dummy model’s behavior. In terms of AUC, all paid LVLMs exhibit relatively higher performance than the dummy model, reflecting stronger discriminative capabilities. In contrast, most free LVLMs perform near the dummy baseline, with four of them achieving even lower AUC scores, indicating a tendency to effectively predict in the wrong direction.

\begin{figure}[t]
    \centering
    \includegraphics[width=1\linewidth]{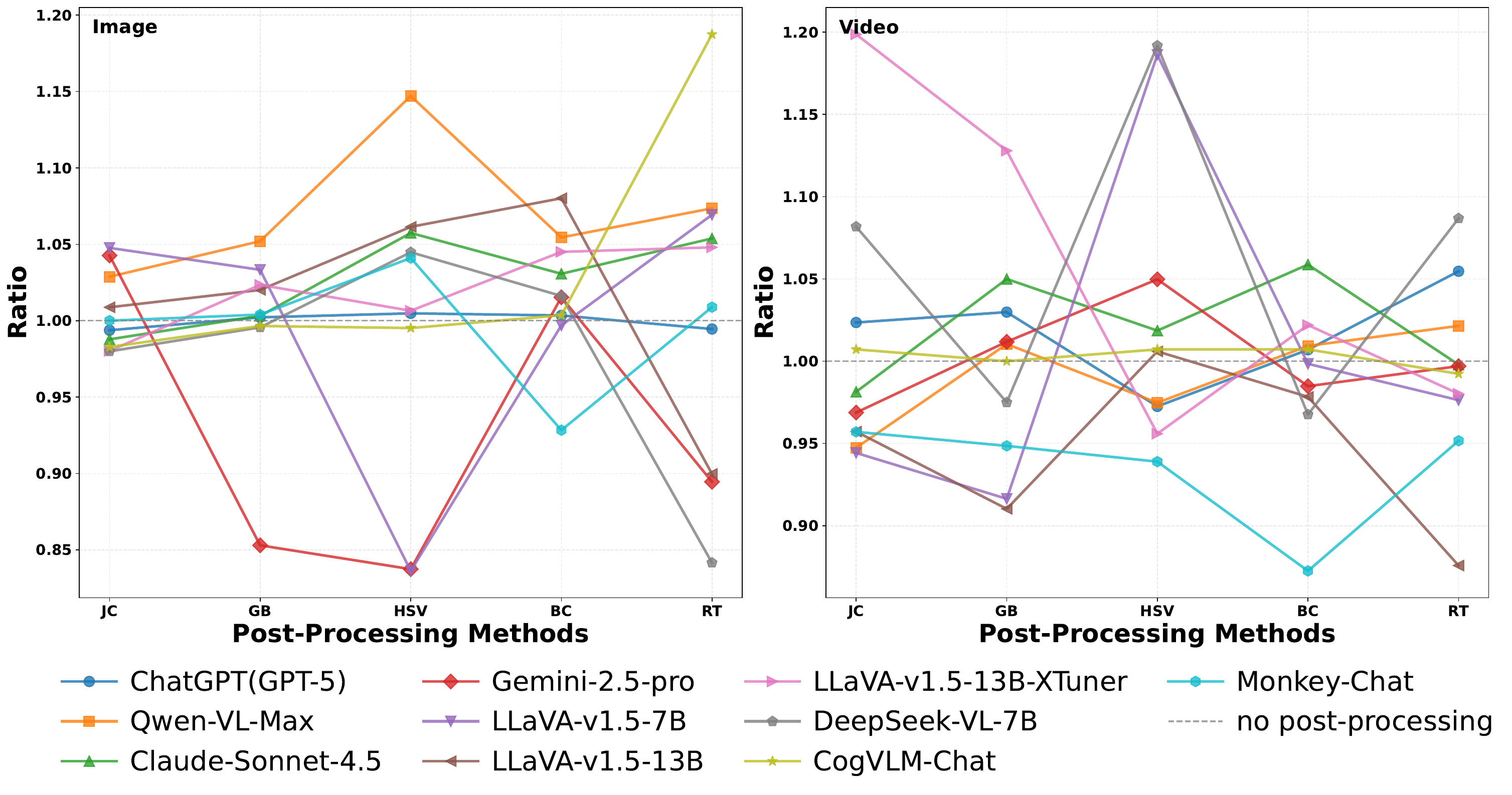}\\[1em] 
    \caption{\emph{Performance (AUC) ratio (original vs. post-processed) for LVLM-aware detectors.}}
    \label{fig:robustness_test}
\end{figure}

As with the robustness evaluation conducted for the LVLM-agnostic white-box detectors, we perform the same set of robustness experiments for all LVLMs. The results, presented in Figure~\ref{fig:robustness_test}, reveal that \textbf{LVLMs exhibit greater robustness compared to detectors from academia and government}, as most models maintain a performance (AUC) ratio within the range of [0.95, 1.05]. A potential explanation is that these LVLMs, such as ChatGPT (GPT-5), are trained on million- to billion-scale multimodal datasets comprising images, videos, and text. Such large-scale and diverse training may enable them to handle variations in input quality more effectively and to internally normalize post-processed inputs to align with the data characteristics encountered during training. However, Qwen-VL-Max demonstrates higher sensitivity to HSV (\ie, random adjustments to Hue, Saturation, and Value channels) post-processing effects on image data, showing a sharp performance decline (around 1.15 ratio). Similarly, several free-access LVLMs (\eg, LLaVA-v1.5-13B-XTuner, LLaVA-v1.5-7B, DeepSeek-VL-7B) experience substantial performance degradation on the video set. Interestingly, we also observe that the performance of certain LVLMs (\eg, Gemini-2.5-pro, Monkey-Chat) actually improves after post-processing (with a lowest ratio close to 0.84), a finding that urges caution rather than confidence, as these transformations should not increase model performance, should remain stable (ratio close to 1.00) if detection techniques are robust. A plausible explanation is that these transformations introduce additional visual artifacts into fake images, making the manipulations more detectable. In cases where these models initially misclassified subtle fakes, the added distortions may have amplified cues that align better with their learned representations of synthetic content, thus enhancing detection accuracy.
These results indicate that, despite their overall resilience, robustness remains a persistent challenge for a few LVLMs.

\section{Discussion}\label{sec:discussion}
Our benchmark presents the first systematic evaluation of existing deepfake detectors on real-world, politically-oriented deepfakes--as well as cheapfakes and alleged instances of false content--collected from social media. Importantly, we adopt the perspective of an average end user, testing each model in its original released form, without additional fine-tuning or domain adaptation, to reflect realistic deployment conditions. According to the experimental results, we have the following findings.

\textbf{Finding 1}: Our analysis of the PDID dataset highlights its \textbf{unique characteristics} and provides insight into the distribution of political deepfakes observed in real-world social media settings in recent years. Compared with existing datasets (Table~\ref{tab:datasets}), the PDID more accurately represents in-the-wild political deepfakes (both images and videos) and provides a valuable foundation for benchmarking detection systems. This fundamentally distinguishes our work from most prior benchmarks (Table~\ref{tab:benchmark-comparison}), which train and evaluate models exclusively on laboratory-generated deepfakes. Such datasets are typically synthesized/generated using a single generative model or tool, limiting their ability to capture the diversity and complexity of real-world manipulations. Moreover, many of these synthetic deepfakes were never publicly disseminated on social media, reducing their practical relevance (\eg, real impacts on the public, elections, or institutional trust) and, consequently, the significance of benchmark results built upon them.


\textbf{Finding 2}: Through frequency-domain analysis of real and fake samples within the PDID, we observe that the authentic media in the PDID differs substantially from the authentic media commonly used to generate laboratory-based deepfakes. This divergence becomes even more pronounced when comparing fake media in the PDID with lab-generated deepfakes. These findings demonstrate the distinctiveness of the PDID dataset, whose \textbf{data characteristics are rarely seen in existing benchmarks}, despite the significance of studying content that has actually been disseminated and interpreted in contexts with potential political implications. The results also suggest that the generative methods underlying the political deepfakes in the PDID may involve previously unseen or heterogeneous synthesis pipelines (\ie, unrestricted to any single model or framework), reflecting the uncontrolled and evolving nature of real-world generative media.

\textbf{Finding 3}: Our evaluation of LVLM-agnostic white-box deepfake detectors indicates that detectors developed in academia and government are \textbf{not yet suitable} for direct deployment in real-world environments. In particular, government-developed detectors exhibit notably low performance, likely due to restrictions in training data and the lack of continued maintenance or updates. This finding underscores the importance of sustained governmental support for deepfake detection research to ensure that publicly funded or endorsed systems remain trustworthy and effective for general users. Fortunately, the mission of the original SemaFor program is being continued under the Digital Safety Research Institute (DSRI) of UL Research Institutes \cite{semaforcontent}, highlighting an ongoing institutional commitment to advancing media forensics. Our results also suggest that such detectors require further fine-tuning and domain adaptation before being applied to politically-oriented or real-world datasets. Moreover, the relatively stronger performance of frequency-based detectors points to a promising direction for improving the generalization ability of LVLM-agnostic white-box models—specifically, by emphasizing the learning of signal-level frequency-domain artifacts during training.

\textbf{Finding 4}: Through our evaluation of LVLM-agnostic black-box deepfake detectors (\ie, commercial tools) from industry, we observe that a few tools do achieve relatively strong AUC performance on the image set. However, most tools exhibit low accuracy and a high FAR, indicating that \textbf{accurate decision-making depends heavily on careful threshold selection}. This suggests that commercial platforms should provide clearer, more intuitive guidance to help non-expert users interpret results effectively to maintain user trust and engagement, though helping users calibrate is far from straightforward and represents an important research challenge. Next, although several tools perform reasonably well on images, they demonstrate a significant drop in performance on videos, revealing that political video deepfake detection remains a major challenge for current commercial tools. Consequently, further development is needed to enhance robustness in dynamic visual contexts. This is critically significant because of the diversity of multimodal content: video with altered frames, authentic videos with authentic audio, authentic images with misleading text captions, and so on. Until such improvements are achieved and tools are carefully, contextually implemented, average users should exercise significant caution when interpreting results from commercial tools, particularly when testing political video deepfakes.

\textbf{Finding 5}: Recently, the use of LVLMs by the general public has increased substantially across a wide range of tasks. Leveraging LVLMs for political deepfake detection has therefore become a promising direction. Our experimental results indicate that \textbf{paid LVLMs relatively outperform free-access counterparts}. This advantage can likely be attributed to two key factors: (1) paid LVLMs are regularly updated and fine-tuned with new and diverse data, improving their adaptability and alignment with recent real-world content—similar to the continuous maintenance in commercial detection tools; and (2) many paid LVLMs incorporate a RAG module \cite{gao2023retrieval}, which enhances performance through online information retrieval and implicit fact-checking. This built-in retrieval capability allows the models to cross-validate visual or textual inputs against up-to-date external knowledge, thereby improving their accuracy in identifying political deepfakes. While the paid LVLMs demonstrate strong potential, achieving high AUC scores in both image and video detection, their overall reliability remains limited due to consistently high FARs. This indicates that users must carefully adjust or interpret probability thresholds before making final authenticity judgments. Nevertheless, one notable advantage of LVLM-based detection is their robustness: \textbf{most LVLMs exhibit stable performance and are largely insensitive to common post-processing operations} applied to the original media, suggesting better resilience to real-world variations in content quality.

\begin{figure}[t]
    \centering
    \includegraphics[width=1\textwidth]{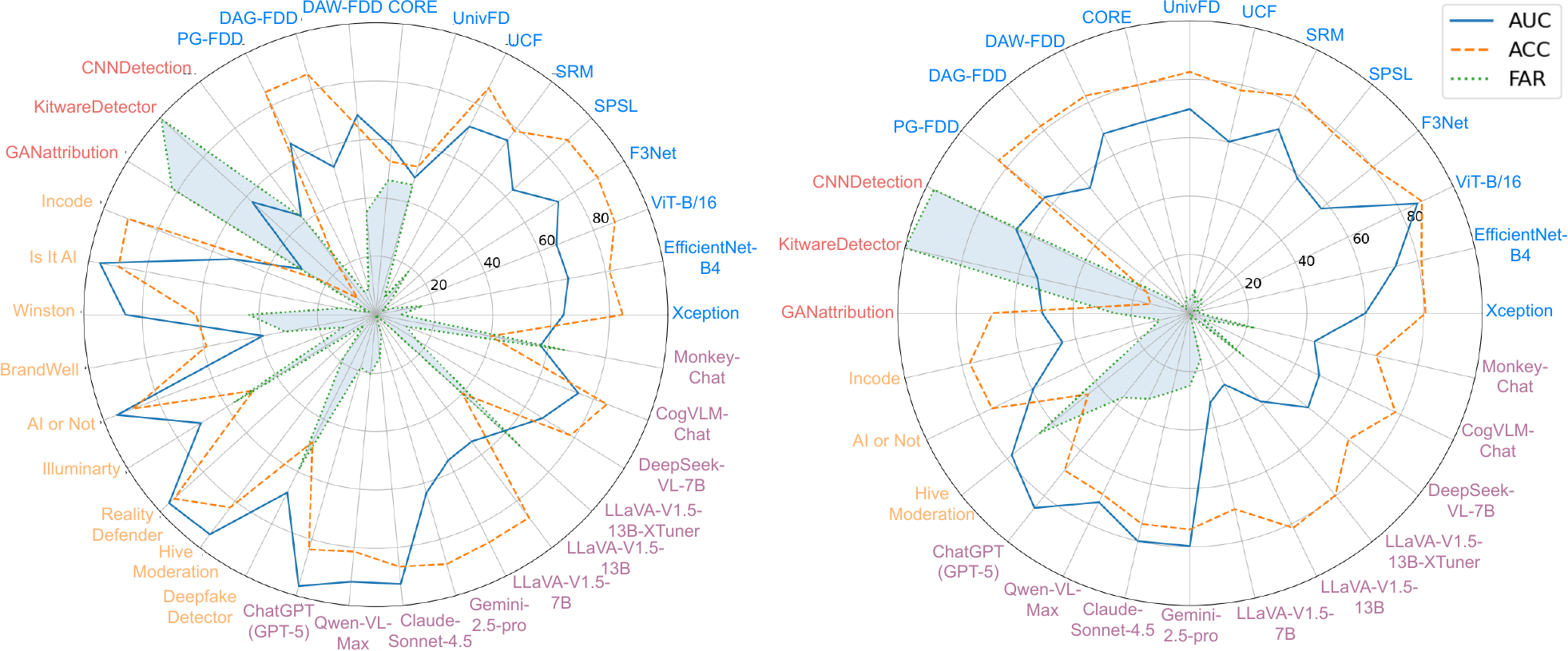}
    \caption{\emph{Performance overview of all detectors on images (Left) and videos (Right).}} 
    \label{fig:radar}
\end{figure}

\textbf{Finding 6}: To better visualize and interpret the performance of current detectors in identifying political deepfakes, we generate radar plots for both image and video sets, as shown in Figure~\ref{fig:radar}. Overall, it is evident that \textbf{paid detectors consistently outperform free-access detectors in terms of AUC}. This discrepancy can likely be attributed to the continued maintenance and regular updates of paid models, which often incorporate newer and more diverse training data, thereby enhancing their generalization capability. However, \textbf{political video deepfake detection remains a major challenge for most detectors}, as reflected by the noticeable performance decline compared to image-level AUC results. Furthermore, our analysis suggests that with careful threshold tuning, it is possible to achieve both high ACC and low FAR, as observed in some academic video detectors. Nonetheless, this process requires substantial technical knowledge and contextual decision-making, which \textbf{limits its practicality for non-expert users} in real-world scenarios.

In general, our systematic benchmarking reveals that existing deepfake detectors struggle to generalize effectively to real-world political cases--the very cases most likely to promote propaganda, distort democratic processes, and undermine trust in news and political institutions. Most detectors exhibit substantial drops in accuracy and AUC when tested on authentic political deepfakes than their original reported performance, potentially due to domain shifts, limited diversity in training data, and overreliance on laboratory-generated synthetic media. Even advanced LVLMs, while more robust to post-processed media, show inconsistent decision thresholds and high false acceptance rates that could mislead average users. These findings reveal an \textbf{urgent need to rethink current detection paradigms}. Future research could prioritize politically contextualized and diversified datasets, cross-modal reasoning, and explainable detection frameworks capable of operating under dynamic real-world conditions. Building detectors that not only identify manipulation but also communicate uncertainty and reasoning transparently will be critical for restoring public trust and ensuring accountability in the age of AI-based generative political media.

Complicating this is the difficulty in calibrating detection tools, communicating appropriate and inappropriate uses clearly, and encouraging contextual adaptation (\eg, by platforms) beyond straightforward integration or binary prediction determinations. Deepfake detection needs to remain accessible from a user design perspective, not just technically robust, or else it risks undermining its own goals and providing false assurance. For instance, a detection tool that only operates on human faces will fail to capture deepfake content of a stolen ballot box, while a detection tool that promises accuracy while failing to address simple manipulations like cheapfakes risks falsely assuring the public, while leaving open major threat vectors for hostile actors to exploit. In sum, despite promising advances made by deepfake detection tools, these tools must be developed, evaluated, implemented, and communicated in the context of real-world problems if they are to deliver on their promise.

Moreover, our results also reiterate the need to approach detection as one layer in a multi-pronged effort to improve public discernment and safeguard platforms. Even effective use of technical detection tools requires, as suggested above, careful communication, user education, and contextual implementation. Beyond that, our study's identification of ground truth labels required a sociotechnical effort--not just examining technical artifacts, but necessarily relying on experts who analyzed source cues and esoteric political context. This strongly suggests that appropriate safeguards require going far beyond technical detection. Media and digital literacy, human-centered design, platform policies, formal regulation, and other layers of governance are required to realize and complement technical detection efforts, if they are to be ultimately successful.

Although our benchmarking study is systematic, several limitations remain. (1) \textit{Limited sample size}. Due to the nature of real-world political deepfakes, collecting a large number of verified samples is extremely difficult, despite our continued efforts through the PDID. Consequently, the relatively small test set may introduce minor statistical bias in the results. Moreover, the majority of deepfakes come from the US political context, limiting both representativeness of the content and generalizability of the evaluation (e.g., a large portion of the content focuses on just a few political figures from a few ethnic backgrounds)  (2) \textit{Focus on visual deepfakes}. Our benchmark exclusively targets visual deepfakes. However, many political manipulations are multimodal, such as those involving synthetic or manipulated audio. While such cases were excluded during data preparation, our findings may not directly extend to more complex multimodal deepfakes that combine visual and audio manipulations. (3) \textit{Face-centered training bias in academic detectors}. The academic detectors included in this study were primarily trained on facial images. As a result, manipulations that occur outside the face region may go undetected if the facial region remains unaltered, implicating the appropriateness of our testing and scoring approach and urging against simplistic interpretations of the results. (4) \textit{Temporal limitation of commercial tool evaluation}. Our evaluation of commercial tools was conducted within a short time window. Given their frequent updates, the performance of individual tools may fluctuate over time. Moreover, due to budget constraints and high subscription costs\footnote{Evaluation procedures for commercial tools relied on their official APIs and documentation. Some providers offered complimentary credits or trial access for preliminary testing; however, all reported results were generated using independently secured access.},
we were unable to continuously monitor performance changes or conduct a broader suite of robustness assessments using multiple post-processing variations. (5) \textit{Subjectivity in ground truth determinations}. Despite efforts to use the most robust existing dataset and incorporate multiple sources of evidence (professional annotators, political and deepfake experts, source cues from social media, external verification from fact checkers), ground truth determinations are not unimpeachable. For instance, it is not given whether an authentic newscast featuring coverage of a deepfake should be considered true or false from a holistic perspective, and either determination could be `unfair' from the perspective of a certain detection procedure or tool. (6) \textit{Scope of evaluated LVLMs}. The LVLMs included in our benchmark were not specifically designed for media forensics. Although deepfake-specialized LVLMs~\cite{lin2024detecting} may yield stronger results, our study intentionally focuses on widely accessible, general-purpose LVLMs to reflect the realistic experience of average users. (7) \textit{Unified prompting strategy}. To maintain consistency across LVLMs, we used a single, standardized prompt. Testing with multiple or adaptive prompts could potentially improve prediction quality but would also introduce greater variability, increase experimental complexity, and compromise comparability across models.

Despite these limitations, we believe that \textbf{our findings are already valuable in revealing the existing challenges and limitations} faced by current detectors in identifying real-world and potentially consequential political deepfakes. The insights gained from this study also highlight several promising directions for future research. In the future, we plan to expand the benchmark by increasing the test sample size and incorporating multimodal deepfakes. Additionally, we intend to evaluate newly developed detectors and deepfake-specific LVLMs to obtain deeper and more comprehensive insights into their effectiveness and generalization across diverse real-world scenarios.



\section{Methods}\label{sec:method}
\begin{figure}[t!]
    \centering
        \includegraphics[width=1\linewidth]{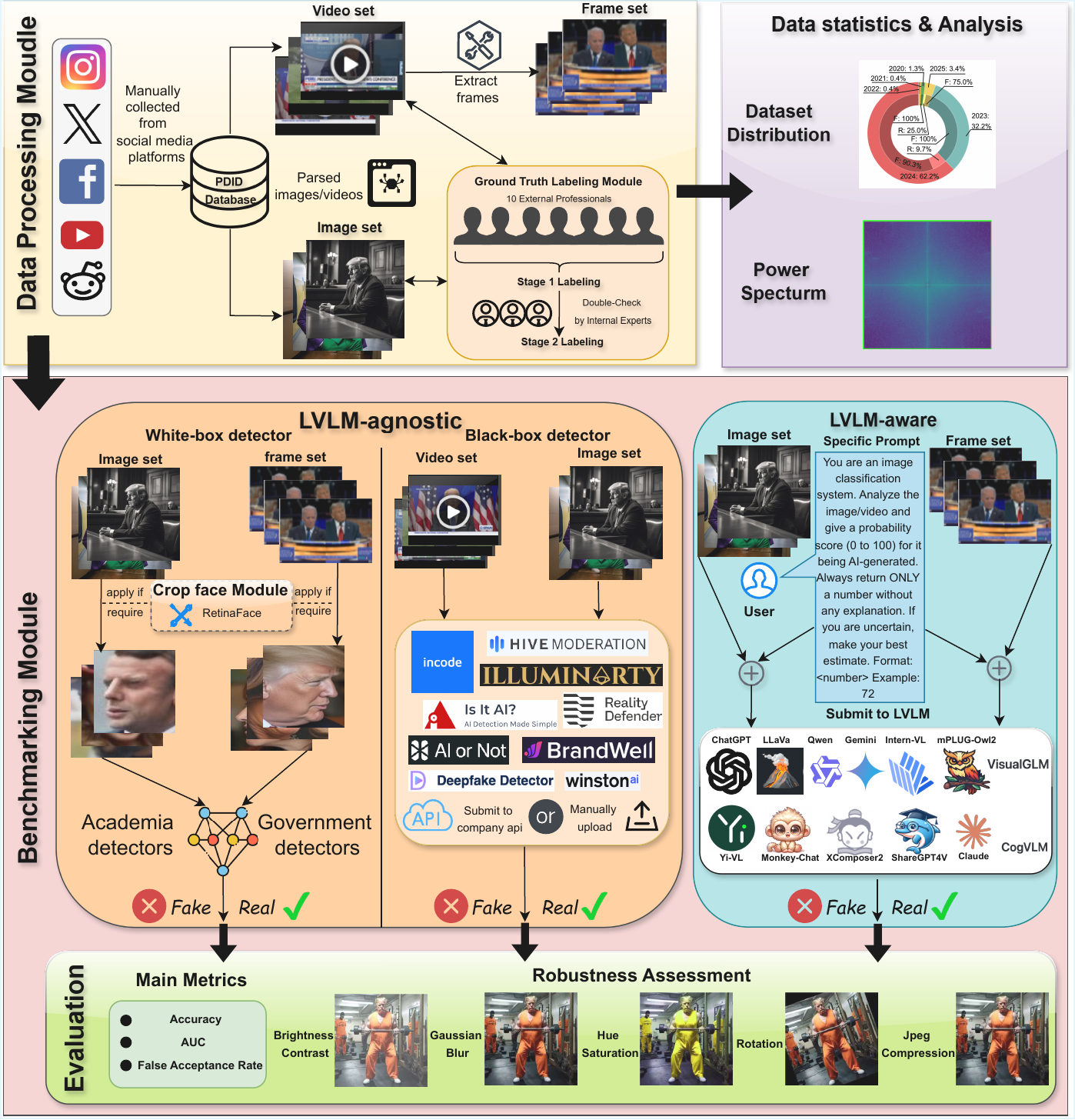}
        \caption{\emph{Overview of the entire pipeline for benchmarking.}}
    \label{fig:pipeline}
\end{figure}
In this section, we describe in detail the methods used for data processing and benchmarking of existing detectors. An overview of the entire pipeline is illustrated in Figure~\ref{fig:pipeline}.

\subsection{PDID}\label{sec:PDID}

\subsubsection{Data Collection}
The PDID provides a novel repository of politically oriented synthesized media. The dataset is inclusive of deepfakes, cheapfakes, and other incidents \textit{alleged} to be deepfakes or cheapfakes. As described in \cite{walker2024merging}, data collection involves a team of individuals-primarily graduate and undergraduate researchers with expertise in political deepfakes-sourcing, ingesting, and annotating identified incidents according to a structured cookbook. Analysts source incidents primarily from five major social media platforms, including monitoring each platform weekly, utilizing independent Democratic-leaning and Republican-leaning accounts, designed to facilitate more robust coverage reflective of the kinds of content that might be targeted at individuals and algorithmically mediated. The data collection process also involves following major news articles: deepfakes reaching this level of attention are de facto considered prominent enough to constitute incidents. Finally, data are ingested from other initiatives aimed at collecting deepfakes, like PolitiFact, Hany Farid's Deepfakes in the 2024 Election Dataset, the WIRED AI Elections Project, and the AI Incidents Database. To improve annotation quality and reliability, incidents are currently primarily limited to English-language and often US-based incidents.

After sourcing, two individuals annotate every incident according to the structured codebook, with reconciliation by a third coder in cases of persistent disagreement. This effort includes capturing an extensive set of metadata, such as media format (image, video, audio-visual), whether the content is \textit{presented} as real or fake, whether the content is determined to be a deepfake or cheapfake, who created or shared the deepfake, metrics on social media views or reposts, and information about who is targeted in the content. We capture incidents that are politically salient deepfakes or cheapfakes, or images or videos that people have called fake, but might be authentic. To be politically salient, the image or video has to either be posted by a politician, include a political figure in the content, or be about a  politically-connected event (e.g., a natural disaster response or election). Additional theoretically informed variables are annotated, covering domains such as communication objectives, depictions of harm, policy narratives, and evidence of real-world consequences. Additionally, evidence of external verification of truth or falsity is captured, focusing on objective information from fact checkers, journalists, or forensic experts, and annotators additionally note both technical artifacts (e.g., too many fingers) and contextual cues (e.g., a certain political figure was no longer living in a given year). Information on self-disclosure (e.g., the poster indicated the image is fake) and on watermarking or labeling is included as well, and annotators incorporate additional qualitative commentary about context and verification to explain incidents and coding decisions and improve research transparency. The complete codebook and associated typologies are available at \href{https://doi.org/10.17605/OSF.IO/FVQG3}{https://doi.org/10.17605/OSF.IO/FVQG3.} 

The initial resulting dataset from which this study's test data set is drawn has a total of 939 images and 502 videos, as of October 2025, covering the time period of 2018 through the present. Unlike purely technical data sets or mere collections of images or videos, the PDID bridges the gap between AI research and sociopolitical analysis by contextualizing incidents within relevant political frameworks, policy narratives, and documented societal impacts, thereby enabling rigorous investigation of synthetic media's role in information integrity challenges. For instance, the rich metadata fields or human explanations provided in the PDID can enable additional research needed to fine-tune reasoning-based LVLMs, drawing from information about source cues, text from broader social media threads, or context about political events that may be difficult to retrieve otherwise, especially retrospectively.



\subsubsection{Two-Stage Label Generation}\label{sec:label-generation}
In this work, we select images and videos from the PDID generated between 2018 and September 2025,  initially excluding those where the research team felt authenticity was too difficult to determine,
resulting in 232 images and 179 videos. While the PDID provides evidence of external verification (e.g., by professional fact checkers), this form of third-party verification only covers a modest subset of images/videos, many of which are labeled as ``unclear either way.'' As such, we engaged in a more intensive \textbf{two-stage} labeling process to identify the ground truth (either \textit{Fake} or \textit{Real}) for all images and videos. 

\textbf{Stage 1 (External Professional Labelers)}:
We first engaged ten professional annotators from an external data labeling company. Since the labelers did not have prior expertise in deepfakes, we provided them with a detailed image and video labeling instruction document (see Supplementary Information for details). To facilitate unbiased human judgment, labelers were not permitted to conduct external research, search online, use fact-checking websites, or employ any automated detection tools. Their decisions were to be based solely on visual and contextual cues described in the provided instructions. Each labeler assigned one of three possible labels to every image or video:
\begin{itemize}
    \item \textit{Real}: Appears authentic, with no clear signs of manipulation.
    \item \textit{Fake}: Appears manipulated, generated, or altered. This includes: deepfakes and AI-generated content, photoshopped images, memes and cartoons, cheapfakes (simple edits), and any content that has been significantly modified from reality.
    \item \textit{Uncertain}: Insufficient evidence to confidently determine authenticity.
\end{itemize}

Specifically, labelers were instructed to consider visual/image cues, social media and contextual cues, real-world plausibility, and self-disclosure indicators when making their determinations. Random guessing was strictly prohibited, and the Uncertain label was to be used only when the labeler was genuinely unable to make a confident judgment. To minimize potential ordering bias, the presentation order of images and videos was randomized for each labeler. After all annotations were completed, we aggregated the results and computed a \textbf{confidence level} for the ``Fake'' label for each sample, defined as follows:
\begin{itemize}
    \item \textit{Low confidence}: Consensus among 0–4 labelers.
    \item \textit{Medium confidence}: Consensus among 5–7 labelers.
    \item \textit{High confidence}: Consensus among 8–10 labelers.
\end{itemize}

\textbf{Stage 2 (Internal Political Deepfake Expert Review)}: Next, a set of researchers with specific expertise in political deepfakes and in the U.S. political context conducted a second round of verification to ensure labeling accuracy. Specifically, we re-examined all images and videos that received \textbf{low} or \textbf{medium confidence} ratings from Stage 1 using a rigorous cross-checking strategy. This included using fact-checking websites such as Snopes \cite{snopes} and AFP fact check \cite{afp} to verify whether images and videos were real or fake, as well as utilizing knowledge about the political context and real-world events, drawing on self-disclosure indicators and information about posters or sharers of content (e.g., usernames, country of origin).
Because the dataset--focused on deepfake incidents--is inherently imbalanced (with a significantly higher number of fake samples than real ones), there is a high risk of mislabeling real images/videos. To mitigate this, the political deepfake expert team additionally reviewed all samples that were labeled as Real by at least one professional labeler in Stage 1. During this stage, we exclude previously labeled real video samples that contain fake audio, as this study focuses exclusively on \textbf{visual deepfakes} and a holistic judgment of truth sensitive to audio would not align with results based on audio-agnostic detection tools (the majority assessed in our study). Based on our expert assessment, we assign final labels. 

It is important to note that many artifacts fell into hybrid categories—for instance, real footage with fake captions (\eg, video footage of a real war, but described as taking place elsewhere), or authentic news segments displaying deepfake clips. In these cases, labeling was based on the dominant communicative intent of the artifact as a whole rather than the presence of localized synthetic elements. This distinction matters for the utility of deepfake detection tools as well as how to fairly evaluate them. Yet in practical detector deployment, providers may overpromise the capabilities of their detectors or underestimate associated challenges, and users likewise may apply such tools to content without such awareness, potentially leading to false confidence. Detection tools that attend narrowly to technical artifacts of diffusion models or GANs risk missing the forest through the trees, as 1) simple edits or manipulations (cheapfakes), 2) strategic hybrid use of authentic and synthetic content or 3) of real and misleading content, or of 4) multi-modal content could induce serious errors in the accuracy of detection tools and impact on the public.

In turn, and to promote a fair evaluation, if a sample was deemed too ambiguous to confidently categorize, we re-designated it as Uncertain and excluded it from the final dataset. Only a small number of samples were excluded in this way. The remaining samples, which achieved high confidence and passed all review criteria, were retained as fake or real. This two-stage validation process resulted in a final curated dataset containing 232 images and 173 videos with verified authenticity labels (\ie, 214 fake images, 18 real images, 146 fake videos, 27 real videos).

\subsubsection{Expose Frequency Domain Artifacts} \label{sec:frequecy}
To identify the specific manipulation artifacts present in the PDID dataset and compare them with those produced by established techniques (\eg, variational autoencoders, generative adversarial networks, and diffusion models), we conducted a frequency domain analysis. This analysis aims to assess whether recent real-world fake images are indeed generated by known academic methods. To this end, we randomly extract one frame from each video using the OpenCV library (\texttt{cv2})~ \cite{opencv_library} and add them to the original image set, resulting in the final test set with 405 (\ie, 232+173) samples used for this experiment. All images and frames are resized to 224$\times$224.

Inspired by \cite{zheng2024breaking}, we extract all fake images. 
For each image $x_i$, where $i\in\{1,\cdots, I\}$ and $I$ is the total number of fake images, we extract its noise residual $r_i$ by subtracting its denoised version. This process is formulated as follows:
\begin{equation} \label{eq:residual}
r_i(m,n) = x_i(m,n) - D\bigl(x_i(m,n)\bigr), \quad i = 1, 2, \ldots, I.
\end{equation}
Here, $(m,n)$ denotes the pixel coordinates, and $D(\cdot)$ represents the denoising filter. In this study, we use the learned denoising network DnCNN~ \cite{corvi2023intriguing} as the denoising filter.

Next, we apply the Discrete Fourier Transform (DFT) to convert each noise residual $r_i$ into its frequency domain representation. Let $M$ and $N$ denote the image height and width, respectively. The transformation is defined as follows:
\begin{equation} \label{eq:dft_alt} 
F_i(k,l) = \sum_{m=1}^{M} \sum_{n=1}^{N} r_i(m,n)\, e^{-j\,2\pi\left(\frac{k}{M}m + \frac{l}{N}n\right)}.
\end{equation}
Here, $(k, l)$ denotes the frequency domain coordinates, and $F_i(k, l)$ is the complex Fourier coefficient at location $(k, l)$ for the $i$-th image.

To capture the characteristic artifacts of the image source, we compute the average power spectral density $S_x(k, l)$ across all images, defined as:
\begin{equation} \label{eq:psd}
S_{\text{PDID}}^{\text{fake}}(k,l) = \frac{1}{I} \sum_{i=1}^{I} \left|F_i(k,l)\right|^2.
\end{equation}
Similarly, we extract all real images and apply the same procedure to obtain $S_{\text{PDID}}^{\text{real}}(k, l)$. 

For comparison, we utilize subsets from the latest lab-based AI-Face dataset~ \cite{lin2025aiface}, which includes traditional deepfake images from four sources, GAN-generated images from ten sources, DM-generated images from eight sources, and their corresponding real images. Specifically, the real images are sourced from FF++ \cite{rossler2019faceforensics++}, DFDC \cite{dolhansky2020deepfake}, DFD~ \cite{google2019deepfake}, and Celeb-DF~ \cite{li2020celeb} for the deepfake category; FFHQ~ \cite{karras2019style} for GANs; and IMDB-WIKI~ \cite{rothe2015dex} for DMs. For each category (\ie, Deepfake, GAN, and DM), we randomly select 1,000 fake images and 1,000 real images from the corresponding source. Following the same procedure described above, we compute $S_{\text{category}}^{\text{fake}}(k, l)$ and $S_{\text{category}}^{\text{real}}(k, l)$, where $\text{category} \in \{\text{Deepfake}, \text{GAN}, \text{DM}\}$. Figure~\ref{fig:power_spectrum} illustrates the spectral artifacts across different categories for both fake and real images.

\subsection{Benchmark Settings}\label{sec:benchmark-settings}
\subsubsection{Detectors}
To build a comprehensive benchmark, we systematically evaluated a diverse set of deepfake detectors, categorized as described below. All detectors were used with their publicly available pre-trained weights or accessed through APIs. No additional training, fine-tuning, or adaptation was performed.
\begin{enumerate}
    \item \textbf{LVLM-agnostic white-box detectors}. In this category, we include detectors developed from academia and government. Specifically, we consider the following models:
    \begin{enumerate}
        \item For detectors developed by the \textbf{academic} community, we adopt 12 pre-trained detectors from the latest benchmark study: AI-Face \cite{lin2025aiface}. These detectors are organized into four major types:
        \begin{enumerate}
            \item \textit{Naive detectors}: Backbone architectures used directly for binary classification, including convolutional models such as Xception \cite{chollet2017xception} and EfficientNet-B4 \cite{tan2019efficientnet}, as well as transformer-based models like ViT-B/16 \cite{vit}.
            \item \textit{Frequency-based Detectors}: Models that exploit frequency-domain cues for forgery detection, such as F3Net \cite{qian2020thinking}, SPSL \cite{liu2021spatial}, and SRM \cite{luo2021generalizing}.
            \item \textit{Spatial-based Detectors}: Models that leverage spatial features (\eg, textures or inconsistencies) for detection, including UCF \cite{yan2023ucf}, UnivFD \cite{ojha2023towards}, and CORE \cite{ni2022core}.
            \item \textit{Fairness-enhanced Detectors}: Approaches designed to improve fairness in AI-generated face detection by incorporating fairness-aware learning objectives. This includes DAW-FDD, DAG-FDD \cite{ju2024improving}, and PG-FDD \cite{lin2024preserving}.
        \end{enumerate}
 
        \item From the \textbf{government} sector, we include three detectors developed under the DARPA Semantic Forensics (SemaFor) program~ \cite{darpa_semafor}, each of which accepts images as input: CNNDetection \cite{wang2020cnn} is a classifier trained on ProGAN \cite{karras2017progressive}-generated images. KitwareDetector \cite{KitwareDetector} is a detector developed by Kitware, Inc., designed to detect GAN-generated images. GANAttribution~ \cite{ganattribution} is a detector that goes beyond binary detection to attribute GAN-generated images to their source generators by identifying distinctive fingerprints left by different generative models. 
    \end{enumerate}
    \item \textbf{LVLM-agnostic black-box detectors}.
    To evaluate the performance of existing commercial deepfake detection tools on our dataset, we first identify nine image-based services and three video-based services. Access to these tools is obtained either through direct purchase, subscription, or by contacting the vendors to request testing access. Specifically, 
    \begin{enumerate}
        \item The image-based commercial tools evaluated are: \textit{Incode, Is It AI, Winston, BrandWell, AI or Not, Illuminarty, Reality Defender, Hive Moderation, and Deepfake Detector}. 
        \item The video-based commercial tools include: \textit{Incode, AI or Not, and Hive Moderation}.
    \end{enumerate}
    It is important to note that a few of these tools reject a subset of the test samples. In such cases, performance metrics are calculated only on the accepted samples for each specific tool.
    
    \item \textbf{LVLM-aware detectors}. Recognizing the rapid progress in LVLMs, we evaluate ten models capable of accepting image inputs and generating textual responses with explanations. Specifically, we assess the performance of 10 LVLMs: ChatGPT(GPT-5) \cite{openai2025gpt5}, Qwen-VL-Max \cite{Qwen-VL}, Claude-Sonnet-4.5 \cite{anthropic2025claude45}, Gemini-2.5-pro \cite{comanici2025gemini}, LLaVA-v1.5-7B \cite{lin2023video}, LLaVA-v1.5-13B \cite{liu2023llava}, LLaVA-v1.5-13B-XTuner \cite{xtuner2023}, DeepSeek-VL-7B \cite{lu2024deepseek}, CogVLM-Chat \cite{wang2024cogvlm}, and Monkey-Chat \cite{Li_2024_CVPR}. For the first four LVLMs, we purchased API access for evaluation. The remaining LVLMs were freely accessible online. To ensure consistent and token-efficient outputs, we used the following prompt for all of them:
    \begin{quote}
    \texttt{You are an image classification system. Analyze the image/video and give a probability score (0 to 100) for it being AI-generated. Always return ONLY a number without any explanation. If you are uncertain, make your best estimate. Format: <number> Example: 72}
    \end{quote}
\end{enumerate}

\subsubsection{Dataset}
In our evaluation, we use the newly constructed image set and video set to assess all detectors.    
To make the comparison fair, we design the following strategies for each detector category.

For \textbf{LVLM-agnostic white-box detectors}, we leverage access to the source code to adapt our data according to each detector's input requirements.
\begin{enumerate}
    \item If a detector expects the full image, we retain the original image format for the image evaluation. For video evaluation, since all detectors in this category are image-based and do not support direct video input, we test the 10 frames extracted for each video. To compute a final prediction score per video, we adopt the following \textit{frame-based aggregation rule}:
        \begin{enumerate}
            \item If any frame yields a detection probability greater than 0.5 (\ie, likely fake), we average only those frame probabilities to obtain the final video-level probability.
            \item If all extracted frame probabilities are below 0.5 (\ie, likely real), we compute the average over all frames.
        \end{enumerate}
    This rule ensures that detectors remain sensitive to subtle manipulations, enabling the identification of a fake video even when only a single frame is manipulated.
    \item If the detector requires cropped facial regions, we apply the RetinaFace model~ \cite{deng2019retinaface,serengil2020lightface} to automatically locate and extract face regions.  The resulting face crops are then resized to the detector-specific input dimensions (\eg, 224$\times$224, 256$\times$256). Note that if no faces are detected in an image or in a whole video, that sample is excluded from the final evaluation.  Therefore,
    \begin{enumerate}
        \item For image evaluation, the above procedure produces a total of 586 cropped face images from the original 232 images. The final prediction score for each original image is defined as the highest face-level prediction score among all detected faces within that image.  
        \item For video evaluation, following the previous rule, we use the 10 frames extracted for each video from the preprocessing procedure. Face cropping is applied where required by the detector. The final prediction score for each frame is defined as the highest face-level prediction score among all detected faces within that frame. The final prediction score for each video is computed according to the \textit{frame-based aggregation rule} described previously.

    \end{enumerate}
\end{enumerate}

For \textbf{LVLM-agnostic black-box detectors}, all models in this category are proprietary and developed by commercial vendors, typically requiring a subscription or one-time purchase for access. We evaluate these detectors using all 232 images and 173 videos. Notably, we do not extract frames from videos for this evaluation, as most detectors in this category return prediction probabilities directly at the video level. 

For \textbf{LVLM-aware detectors}, several models (\eg, ChatGPT, Qwen) require a subscription or purchase for access. To manage cost while maintaining consistency, we evaluate all detectors in this category using the original curated set of 232 images and 173 videos. For video testing, we use the 10 frames extracted for each video. We then follow the same aggregation rule used for LVLM-agnostic white-box detectors: if any frame yields a probability above 0.5 (indicating likely fake), we compute the final video score by averaging those frames; otherwise, we average all extracted frame scores. This approach ensures consistent and cost-effective evaluation across detectors.

\textbf{Robustness Assessment}. Although our media are originally collected from social media, they may be redistributed across various platforms in the future. It is therefore critical to evaluate the performance of existing detectors under such dissemination scenarios. To this end, we conducted a robustness assessment for both LVLM-agnostic black-box detectors and LVLM-aware detectors. Following the protocol in~ \cite{lin2025aiface}, we applied five post-processing operations to all images and extracted video frames to simulate real-world degradations. The transformations include:
\begin{itemize}
    \item JPEG Compression (JC) – re-encoding images at a fixed quality level;
    \item Gaussian Blur (GB) – applied with a predefined kernel size;
    \item HSV Shift (HSV) – random adjustments to hue, saturation, and value channels;
    \item Brightness/Contrast Adjustment (BC) – random modifications within specified bounds;
    \item Rotation (RT) – applied up to a fixed angle, with padding added to preserve image dimensions.
\end{itemize}

\subsubsection{Evaluation Metrics}
To evaluate detection performance, we use the following utility metrics: ACC, AUC, and FAR. Higher values of ACC and AUC indicate better performance, while a lower value of FAR is preferred. Unless otherwise specified, a fixed threshold of 0.5 is applied to compute ACC and FAR. For LVLM-agnostic black-box detectors, if a tool provides binary output (\ie, ``Real" or ``Fake"), we use it directly. If only a probability score is returned, we apply a threshold of 0.5 to determine the final prediction label for each sample.

\section*{Data availability}
The raw PDID database can be found at \url{http://bit.ly/pdid}.
The curated PDID dataset for the benchmarking can be accessed at 
\url{https://purdue0my.sharepoint.com/:f:/g/personal/lin1785_purdue_edu/EpQPBynMPlJDkdQ05Q-Ej5EBNzoiQIhZgFHLo6cbrnYNMw?e=AWkXKE}.

\section*{Code availability} 
All of the code used in this paper is available under the MIT License at \url{https://github.com/Purdue-M2/Political_Deepfakes_Benchmark}.

\section*{Acknowledgements}
This work is supported by the U.S. National Science Foundation (NSF) under grant IIS-2434967 and the National Artificial Intelligence Research Resource (NAIRR) Pilot and TACC Lonestar6. The views, opinions and/or findings expressed are those of the author and should not be interpreted as representing the official views or policies of NSF and NAIRR Pilot.

\section*{Author contributions}
S.H. and D.S.S. conceived the project and designed the study as well as the benchmarking framework.
G.L. collected and curated datasets, conducted experiments, and contributed to data analysis.
C.P.W. assisted with dataset curation and contributed to manuscript drafting.
L.L. performed experiments, analyzed data, and co-drafted the manuscript.
S.H. and D.S.S. provided major revisions, and supervised the overall research.
All authors contributed to the discussion, interpretation of results, and final manuscript preparation.

\section*{Competing interests}
The authors declare no competing interests.


\backmatter


\section*{Declarations}

\begin{itemize}
\item Funding

U.S. National Science Foundation (NSF) under grant IIS-2434967.
\item Conflict of interest/Competing interests (check journal-specific guidelines for which heading to use)

The authors declare that they have no known competing financial interests or personal relationships that could have appeared to influence the work reported in this paper.
\item Ethics approval and consent to participate

Not applicable
\item Consent for publication

Not applicable

\item Materials availability

Not applicable

\end{itemize}








\newpage
\clearpage
\begin{appendix}
\begin{center}
\textbf{\Large Supplementary Information}
\end{center}

\section{Image and Video Labeling Instructions}




You will evaluate a dataset of images and videos (around 400 items). Each item will be reviewed by 100 labelers. Your task is to use only the visual content and the information presented in or around the item itself to decide whether it is real, fake, or uncertain.

You are not expected to conduct outside research or use specialized detection tools.

\subsection{Your Task}
For each image or video, select one label:
\begin{itemize}[leftmargin=*]
  \item \textbf{Verified Real} — Appears authentic, with no clear signs of manipulation.
  \item \textbf{Verified Fake} — Appears manipulated, generated, or altered (including deepfakes, cheapfakes, photoshops, memes, cartoons, or cases where context strongly suggests alteration).
  \item \textbf{Uncertain} — Not enough evidence to confidently judge either way.
\end{itemize}

\subsection{What to Consider}

\subsubsection{Visual/Image Cues}
\begin{itemize}[leftmargin=*]
  \item \textbf{Distortions:} unnatural hands, teeth, or ears; mismatched lighting; warped or duplicated backgrounds.
  \item \textbf{Overlays:} added text, watermarks, or meme elements.
  \item \textbf{Cartoonish style:} clearly non-photographic or drawn.
\end{itemize}

\subsubsection{Social Media and Contextual Cues}
\begin{itemize}[leftmargin=*]
  \item \textbf{Source clues:} verified accounts may increase credibility; parody accounts may signal unreliability. Neither is conclusive on its own.
  \item \textbf{Timing cues:} inconsistent timestamps or dates may raise suspicion.
  \item \textbf{Format clues:} meme layouts, joke captions, or doctored screenshots are signals.
\end{itemize}

\subsubsection{Real-World Plausibility}
\begin{itemize}[leftmargin=*]
  \item \textbf{Event plausibility:} if the depicted situation is highly improbable, it may suggest fakery.
  \item \textbf{Location cues:} unexpected or impossible settings for the subject.
  \item \textbf{Timing plausibility:} if the same person appears in two places simultaneously, it’s likely fake.
\end{itemize}

\subsubsection{Self-Disclosure and Labels}
Explicit labels (e.g., “this is a parody,” “AI-generated”) or usernames like \texttt{@fun-deepfake-editor} are strong signals but not absolute proof. Treat them as evidence to weigh holistically.

\subsection{What NOT to Do}
\begin{itemize}[leftmargin=*]
  \item Do not Google, check news, or verify against external fact-checks.
  \item Do not use technical deepfake detection tools.
  \item Do not guess randomly—if uncertain, select \textbf{Uncertain}.
\end{itemize}

\subsection{Best Practices}
\begin{itemize}[leftmargin=*]
  \item Use all cues together: visuals, context, plausibility, and disclosures.
  \item Stay balanced—no single cue determines the label.
  \item Be consistent—apply the same reasoning across items.
  \item Multiple annotators will review each item; thoughtful, careful judgments are more valuable than quick guesses.
\end{itemize}

\subsection{Examples}
\subsubsection{Verified Fake}
\begin{itemize}[leftmargin=*]
  \item TikTok video watermarked “deepfake parody” with clear facial distortions.
  \item Photoshop of a politician shaking hands with an alien, mismatched lighting.
  \item Meme screenshot with overlaid joke captions.
\end{itemize}

\subsubsection{Verified Real}
\begin{itemize}[leftmargin=*]
  \item Press photo of a candidate at a rally, with no visible anomalies.
  \item Screenshot of a televised news broadcast showing plausible visuals and context.
\end{itemize}

\subsubsection{Uncertain}
\begin{itemize}[leftmargin=*]
  \item Medium-quality video where lip-syncing looks off.
  \item Screenshot from a non-verified account showing plausible but unconfirmed content.
  \item Official post with slight oddities—conflicting signals.
\end{itemize}

\subsection{Frequently Asked Questions (FAQ)}
\begin{itemize}[leftmargin=*]
  \item \textbf{Q1:} What if the content is a meme or cartoon?  
  \textbf{A:} Generally treat as Fake since the image has been altered.

  \item \textbf{Q2:} What if the content comes from a parody or joke account?  
  \textbf{A:} Treat as a signal but not definitive; weigh it against visual cues.

  \item \textbf{Q3:} What if the content is labeled “AI-generated”?  
  \textbf{A:} Treat as strong evidence of fakery but verify visually.

  \item \textbf{Q5:} What if glitches might be due to compression or manipulation?  
  \textbf{A:} Choose Uncertain.

  \item \textbf{Q6:} What if the event seems implausible?  
  \textbf{A:} If extremely implausible, lean Fake; if merely unusual, consider Uncertain.

  \item \textbf{Q7:} What if the item is a screenshot of another post?  
  \textbf{A:} Use both the image and contextual cues to evaluate authenticity.
\end{itemize}

\newpage
\section{Details of Commercial Tools Testing}
All commercial tools are tested through their official APIs; if an API is not available, we manually upload images or videos one by one for evaluation. All 9 tools successfully processed all image and video samples, except for two tools that automatically rejected a subset of inputs due to internal filtering or format restrictions.

\textbf{Incode} is a face-oriented deepfake detector primarily designed for identity verification and biometric authentication. It automatically rejects inputs when no face is detected or when the detected region is too small. 
\textbf{BrandWell} accepts only images in .jpg, .png, or .webp formats. 

\section{Details of LVLM Testing}
Paid LVLMs were evaluated through their official APIs. However, we occasionally observed that the models failed to return responses in the expected format. To maintain experimental consistency, we did not modify the original prompts (although we attempted several variations to elicit valid score outputs, without success). Additionally, due to company-imposed content restrictions, the models sometimes halted generation when the input was deemed inappropriate.

\end{appendix}



\newpage
\clearpage
\printbibliography

\end{document}